\documentclass[sn-mathphys-num]{sn-jnl}

\usepackage{graphicx}%
\usepackage{multirow}%
\usepackage{amsmath,amssymb,amsfonts}%
\usepackage{amsthm}%
\usepackage{mathrsfs}%
\usepackage[title]{appendix}%
\usepackage{xcolor}%
\usepackage{textcomp}%
\usepackage{manyfoot}%
\usepackage{booktabs}%
\usepackage{algorithm}%
\usepackage{algorithmicx}%
\usepackage{algpseudocode}%
\usepackage{listings}%
\usepackage{soul}%

\usepackage{makecell}%
\usepackage{soul}%
\usepackage{color}%
\usepackage{float}%
\usepackage{subcaption}%
\usepackage{chngpage}%
\usepackage{lmodern}%
\usepackage{anyfontsize}%

\theoremstyle{thmstyleone}%
\theoremstyle{thmstyletwo}%

\theoremstyle{thmstylethree}%

\begin{document}

\title[Training Variation of Physically-Informed Deep Learning Models]{Training Variation of Physically-Informed Deep Learning Models}


\author[1,2]{\fnm{Ashley} \sur{Lenau}}\email{atlenau@sandia.gov}

\author[3,1]{\fnm{Dennis} \sur{Dimiduk}}\email{dennis.dimiduk@bluequartz.net}

\author*[1,4]{\fnm{Stephen R.} \sur{Niezgoda}}\email{niezgoda.6@osu.edu}

\affil[1]{\orgdiv{Department of Materials Science and Engineering}, \orgname{The Ohio State University}, \orgaddress{\city{Columbus}, \postcode{43210}, \state{Ohio}, \country{United States}}}

\affil[2]{\orgdiv{Center for Integrated Nanotechnologies}, \orgname{Sandia National Laboratories}, \orgaddress{\city{Albuquerque}, \postcode{87185}, \state{New Mexico}, \country{United States}}}

\affil[3]{\orgname{BlueQuartz Software}, \orgaddress{\city{Springboro}, \postcode{45066}, \state{Ohio}, \country{United States}}}

\affil[4]{\orgdiv{Department of Mechanical and Aerospace Engineering}, \orgname{The Ohio State University}, \orgaddress{\city{Columbus}, \postcode{43210}, \state{Ohio}, \country{United States}}}


\abstract{A successful deep learning network is highly dependent not only on the training dataset, but the training algorithm used to condition the network for a given task. The loss function, dataset, and tuning of hyperparameters all play an essential role in training a network, yet there is not much discussion on the reliability or reproducibility of a training algorithm. With the rise in popularity of physics-informed loss functions, this raises the question of how reliable one's loss function is in conditioning a network to enforce a particular boundary condition. Reporting the model variation is needed to assess a loss function's ability to consistently train a network to obey a given boundary condition, and provides a fairer comparison among different methods. In this work, a Pix2Pix network predicting the stress fields of high elastic contrast composites is used as a case study. Several different loss functions enforcing stress equilibrium are implemented, with each displaying different levels of variation in convergence, accuracy, and enforcing stress equilibrium across many training sessions. Suggested practices in reporting model variation are also shared.}




\maketitle

\section{Introduction}\label{sec1}
Repeatable experiments are the cornerstone of the scientific method. They solidify our understanding and allow scientists to improve theory, apparatus, analysis tools, simulations, etc. A lack of reproducibility in an experiment can be a sign of a flawed understanding of the principles behind the experiment or a flaw in the design of the experiment itself. Materials-based direct numerical simulations, on the other hand, are derived from theory and are typically deterministic - thus essentially 100\% repeatable \footnote{Of course stochastic simulations are important for analyzing the performance of systems whose
behavior depends on the interaction of random processes, typically described by probability models. However, the discussion of stochastic simulation is outside the scope of this work.}. Error in a computational model is typically a result of model form error, incomplete or incorrect physics in the model, or parameter uncertainty, which is often associated with uncertainty in material properties. Computational uncertainty quantification involves characterizing these uncertainties through the comparison of simulation results with experimental data as part of a Verification and Validation process. The first step to quantifying the uncertainty of a simulation method is understanding the systematic variance and its sources, which will provide information on a method's reproducibility. In this paper, variance is of particular interest in machine learning (ML) models where different random initializations could lead to different levels of performance for the same method.

In recent years, ML models have seen a rise in popularity in computational materials science. ML and deep learning (DL) models have been used for experimental analysis \cite{DL_micro,JACOBS2022}, materials discovery \cite{Merchant2023,ZUO2021}, material property prediction \cite{LI2019,jha2022}, and microstructure generation \cite{HENKES2022,chun2020}, to name a few. Despite these advancements, ML has faced criticism for its ``black box'' nature in that many question what the model is learning during the training phase \cite{Agrawal_Choudhary_2019}. While ML models used in materials science are typically deterministic during execution, meaning that the same inputs will produce the same outputs, modern strategies for ML training utilize stochastic sampling of the training data and parameter initialization, resulting in different network behavior from one training cycle to another. Predictions of ML models are non-deterministic because of the many random variables that go into training and initializing the model, making the variation and repeatability of these models a concern in scientific applications.

The repeatability of training a ML model is an active area of research, and several publications have outlined the many variables that can cause variability in model performance. Ref. \cite{pham2021} implemented several networks with identical training and showed that performance can vary as much as 10.8\%. They also show that a network trained 16 times with the same random seed can have drastically different convergence times, as large as a 145.3\% relative difference. Ref. \cite{khan2019} showed that different classifiers had different levels of variation in performance for different datasets. Ref. \cite{bouthillier2021} showed that different hyperparameter optimization and learning procedures lead to different variations in performance for different networks. Ref. \cite{alahmari2020} initialized a network with the same random seed and implemented the network using different versions of TensorFlow, resulting in different accuracy across several training sessions. Even networks trained with a fixed random seed can have variances in biases from different training sessions, as demonstrated by Ref. \cite{qian2021}. The precision used to train the network (double versus single precision) also affects variability, with Ref. \cite{Pinto_Alguacil_Bauerheim_2022} showing that double precision leads to more uniform predictions. Having all these factors causing model performance variations, there is a need to provide code, check for reproducibility through re-implementation \cite{Pineau}, and also report a model's variation. Reporting a network's improvement compared to another implementation is statistically meaningless unless the model performance variation is also reported, as noted by Pham et al. (2021). For example, Pham and co-authors showed that a network reporting a 0.8\% increase in accuracy had a variation in accuracy of 2.9\%, meaning that the ``improvement'' is within the bounds of accuracy variability. 

Reporting the random seed initialization used for training (as suggested by \cite{best_prac_sparks}'s best practice guide for material scientists) may ensure reproducibility for a single random seed, model architecture, and training strategy, but does not evaluate the consistency of a model architecture or training strategy. This is particularly important when comparing different ML methods for things such as convergence rate or prediction error. Training with multiple random initializations evaluates the reliability and reproducibility of a training strategy. Without this evaluation, the robustness of a training strategy in reference to random seed initialization sensitivity, or slight changes in datasets will remain completely unknown. For example, deep material networks (DMNs) \cite{shin_dmn} train to homogenize effective material properties for a single composite microstructure, allowing the user to quickly iterate through many different phase properties to achieve a desired effective composite property. While DMNs are fairly cheap to train, they require to be re-trained if the geometry of the composite is changed. For ML models such as DMNs that often need to be re-trained, evaluating the robustness and consistency of a training strategy is extremely important.

In general, there have been limited efforts in studying or reducing a ML model's training variability. More effort has been focused on defining the epistemic uncertainty of ML model predictions \cite{tavazza2021,OLIVIER2021,Viana2021,LI2024} or reducing the variability of predictions of a model from a single training run \cite{lemay2022improving}. Other studies have shown that networks with physically-informed losses have lower standard deviations in prediction errors \cite{BAI2023,LU2022,Krishnapriyan2021}, with Ref. \cite{Krishnapriyan2021} demonstrating that a curriculum training strategy reduces performance variation of physics-informed neural networks (PINNs).
Adding physics-informed losses ensures that the ML model is learning a specified physical boundary condition, adding some interpretability to the training process. Developing ML models for scientific applications require careful and fair comparisons of current training methods and loss functions to previous ones, and we argue that training consistency and reproducibility should be carefully considered in these model comparisons. 
Yet, recent review articles focused on physics informed ML show a significant lack of discussion on the reproducibility of physics informed ML models \cite{meng_rev_2025,cuomo_rev2022}, with \cite{cuomo_rev2022} identifying the initialization of PINN models a future research direction. 
In the context of physics informed ML, very few studies report the training variability in any context.
Ref. \cite{b_and_e} showed that from a sample of ten different training initializations a physics-informed neural network typically had less success in training the eikonal equation with rectified linear unit (ReLU) activation than with leaky ReLU activation.
Ref. \cite{pang_fpinn} evaluated their fractional physics-informed neural network's sensitivity to parameter initializations by reporting the standard deviation of the network's error across ten different trainings.

Measuring a training method's variability along with its accuracy is not only important in ML model development but in computational materials science as well where physics-informed losses are prominent and often used to comparatively reduce errors, data requirements, convergence, etc. However, as outlined in the previous paragraphs, physically-informed training strategies' effect on training variability is not well defined in literature, and any comparative improvements by using a physics-informed loss may be in question if the training variability is not reported.

Previous work \cite{lenau2024importance} describes a Pix2Pix network \cite{isola2018} that predicts the stress fields of a high elastic contrast two-phase composite. Several loss functions having physics-based regularization (PBR) were implemented to enforce stress equilibrium and their performance was compared to a network without PBR. Each implementation was trained ten times, and the stress and equilibrium errors were averaged across the different training sessions. We showed that the PBR losses reduced the equilibrium mean squared error ($\text{MSE}_{equil}$) by at least 51\%, with two of the PBR losses maintaining similar stress errors compared to the network without PBR.  However, when comparing the best-performing models of the different training sessions for each method, the baseline network's best-performing model outperformed the other PBR models' stress errors by 4.5\%, even though another model with PBR outperformed the baseline on average. This raised the question of how each method varied in performance across different training sessions and if perhaps PBR methods reduce this variation. 

In this work, the variability across different training sessions of the three different PBR methods from the previous work is evaluated and compared to the baseline model without PBR. 
The current study will focus on the consistency of model predictions and errors
to assess the performance and reliability of each implementation as a training strategy for stress field prediction. Several PBR methods are implemented to generalize their effect on training a model in comparison to the baseline model. Different feature representation across the different methods and trainings within a method are additionally investigated.

\section{Methods}

A Pix2Pix generative adversarial network (GAN) \cite{isola2018} was used to predict the normalized stress fields of a high elastic contrast two-phase composite. The network takes in the spatial arrangement of the phases in the composite as input and predicts the corresponding 2D stress fields as output. Several different PBR terms were implemented to penalize generated stress fields' deviation from stress equilibrium through the evaluation of stress divergence ($\nabla\cdot\sigma=0$, neglecting external body forces). A complete description of the implemented PBR strategies can be found in \cite{lenau2024importance}. The following are brief summaries for each PBR method. 

\begin{description}
    \item[Simple Addition Regularization]  The weighted addition of the absolute stress divergence values directly to the objective function, analogously to the L1 regularization implemented in the base Pix2Pix network. This method biases the Generator to synthesize stress fields that are closer to equilibrium. (see Equation 4 in \cite{lenau2024importance})
    \\
    \item[Sigmoid Regularization] The root-mean-square of a divergence field is evaluated to encourage predictions to have similar error convergence as the training data. This method biases the discriminator with an additional weight that captures the probability of whether a divergence field is calculated from a set of “real” (from the training dataset) or synthetic stress fields. (see Equations 5-8 in \cite{lenau2024importance})
    \\
\item[$\mathit{\tan^{-1}}$ Regularization] This method encourages the Generator network to produce stress fields that have similar divergence errors to the training dataset. The $\tan^{-1}$ function is chosen to stabilize gradients of the loss function at large errors while also decreasing to zero in the limit of small deviations in the divergence between generated and training stress fields. (see Equations 9-10 in \cite{lenau2024importance})
\end{description}

\noindent These models were compared to a network that uses the original Pix2Pix objective without any PBR terms, referred to as ``baseline'' (see Equation 2 in \cite{lenau2024importance}). 

The same spinodal decomposition (with varying phase field parameters) dataset, network architecture, regularization methods, and their hyperparameters were used as in our previous work. The various networks were trained on networks trained on identical datasets containing 1,025 image pairs. In this work, each implementation was trained 30 times on the same dataset. Previous work trained each network 10 times and this work trains each network an additional 20 times, for a total of 30 training sessions. A random seed was not fixed for any of the training sessions. 
A validation dataset was used to evaluate hyperparameters and model selection (described in previous work \cite{lenau2024importance}). Checkpoints were saved during training every 1,000 iterations and were used to evaluate the training statistics after training with a test dataset containing 205 image pairs.

\begin{figure}[]
\centering
\captionsetup{width=\linewidth}
\begin{subfigure}{0.4\textwidth}
    \centering
        \hspace*{-1.1in}
        \includegraphics[width=\linewidth]{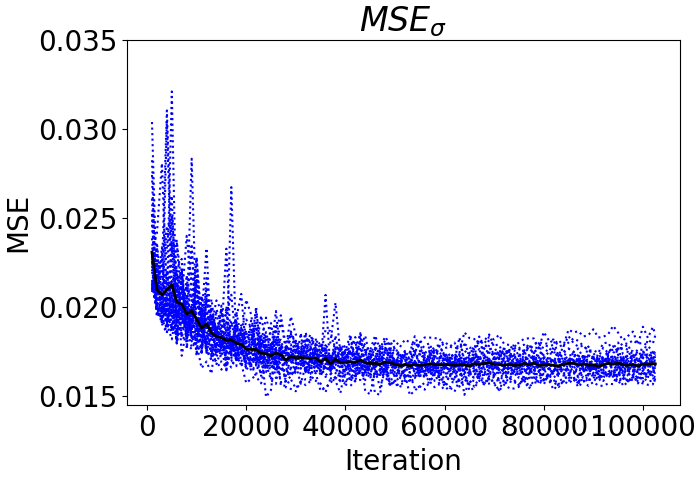}%
        \includegraphics[width=\linewidth]{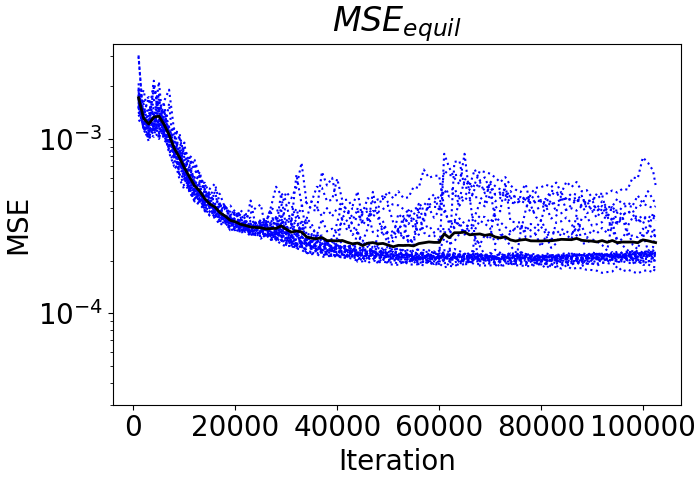}
        \caption{Baseline}
        \label{graph_tuning_NR}
\end{subfigure}

\begin{subfigure}{0.4\textwidth}
    \centering
        \hspace*{-1.1in}
        \includegraphics[width=\linewidth]{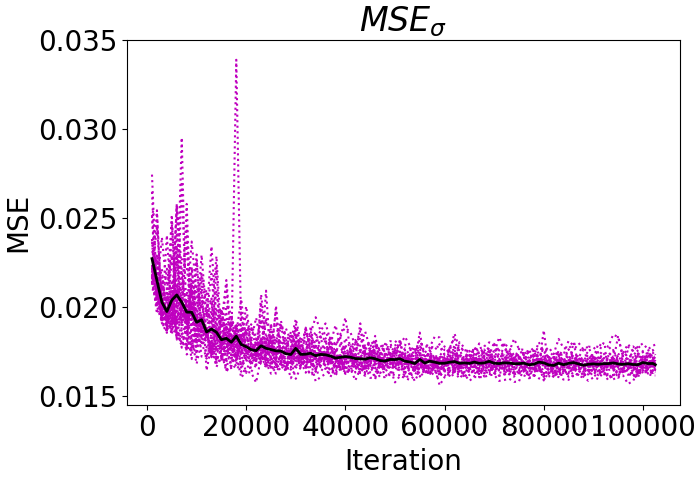}%
        \includegraphics[width=\linewidth]{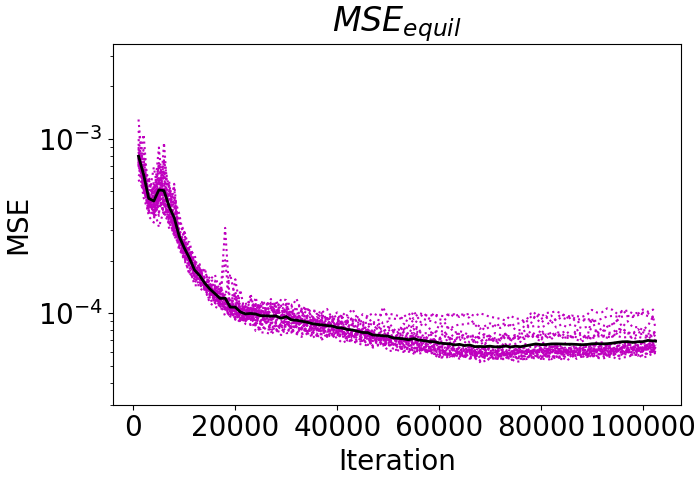}
        \caption{Sigmoid}
        \label{graph_tuning_Sig}
\end{subfigure}

\begin{subfigure}{0.4\textwidth}
    \centering
        \hspace*{-1.1in}
        \includegraphics[width=\linewidth]{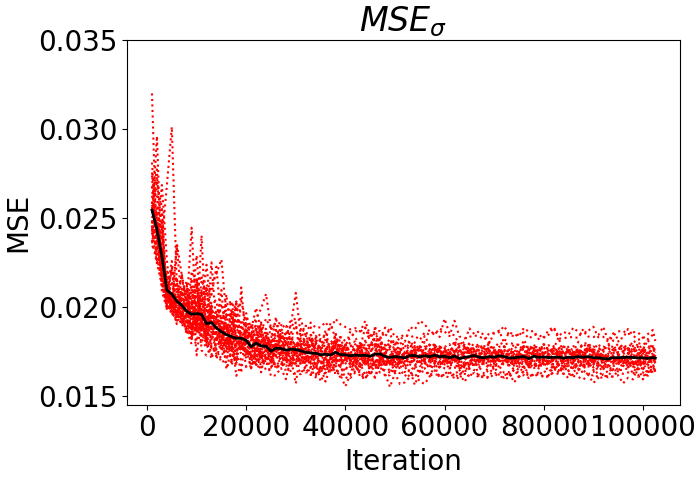}%
        \includegraphics[width=\linewidth]{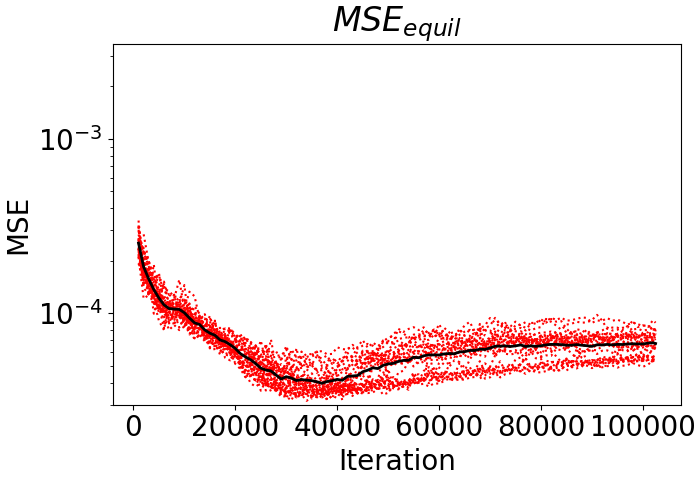}
        \caption{Simple Addition}
        \label{graph_tuning_AD}
\end{subfigure}

\begin{subfigure}{0.4\textwidth}
    \centering
        \hspace*{-1.1in}
        \includegraphics[width=\linewidth]{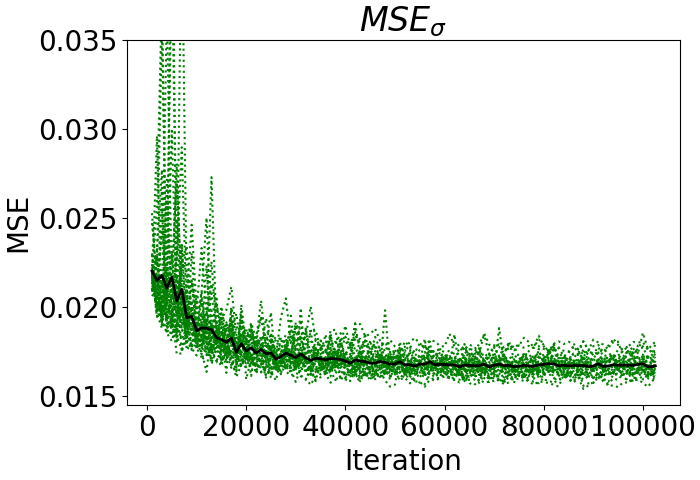}%
        \includegraphics[width=\linewidth]{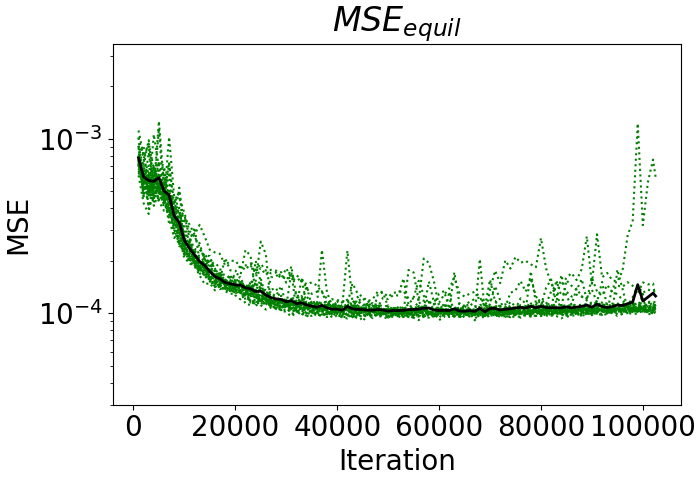}
        \caption{$Tan^{-1}$}
        \label{graph_tuning_TN}
\end{subfigure}

\caption{The $\text{MSE}_{\sigma}$ (left) and $\text{MSE}_{equil}$ (right) throughout training for each of the 30 training sessions. The solid black line indicates the average across the 30 different sessions. (a) No physics-based regularization, (b) simple addition, (c) sigmoid, and (d) $tan^{-1}$.}
\label{fig:training}
\end{figure}

\section{Results}

Mean squared error (MSE) of the stress fields ($\text{MSE}_{\sigma}$) is used to generalize how accurately the model is predicting the values in the stress fields, while the MSE of divergence ($\text{MSE}_{equil}$) summarizes how well the model satisfies stress equilibrium in the predicted stress fields. It should be noted that the $\text{MSE}_{equil}$ in predictions is in comparison to the elasto-viscoplastic crystal-plasticity fast Fourier transform (CP-FFT) simulations used to generate the training and validation datasets, not from zero. The CP-FFT simulation converges to stress equilibrium iteratively, with the average root-mean-square of stress divergence for a set of stress fields being $\sim10^{-3}$ for the dataset.

$\text{MSE}_{\sigma}$ and $\text{MSE}_{equil}$ are shown in Figure \ref{fig:training} for each method throughout training for each of the 30 different training sessions. The predicted stress field errors are very similar between each method, which was also observed in the previous work. The $tan^{-1}$ method has greater $\text{MSE}_{\sigma}$ values earlier on in training (up to about 0.045), but still converges to similar $\text{MSE}_{\sigma}$ values as the other methods. The baseline method has visibly more variation in $\text{MSE}_{equil}$ than the models having PBR. The $tan^{-1}$ model has more variation than the other PBR models, but still less than the baseline model. The PBR methods also significantly reduce $\text{MSE}_{equil}$ compared to the baseline model. 

Table \ref{tab:avg_perform} shows the ``average performance'' of each method. The average performance describes the average errors of the best-performing iterations across the 30 training sessions for a given method. The best-performing iteration is defined as the lowest average $\text{MSE}_{\sigma}$ from the test dataset and is found for each training session. The average iteration at which this occurs is reported in Table \ref{tab:avg_perform} for each method. The deviations in Table \ref{tab:avg_perform} show how much the average error and iteration will vary across different training sessions. On average, the sigmoid method will need a little more time to reduce $\text{MSE}_{\sigma}$ than the baseline method, while the simple addition and $tan^{-1}$ methods will need less time than the baseline method. The sigmoid and simple addition methods are likely to have slightly higher errors in their stress fields for a given training session, but significantly lower errors in equilibrium than the baseline and $tan^{-1}$ methods. A similar trade-off between optimizing $\text{MSE}_{\sigma}$ and $\text{MSE}_{equil}$ is seen in the previous work. The average $\text{MSE}_{\sigma}$ deviates less in the sigmoid and $tan^{-1}$ methods and all PBR methods have less deviation in $\text{MSE}_{equil}$ than the baseline model. The errors from Table \ref{tab:avg_perform} are normalized to the baseline model in Table \ref{tab:norm_perf} for easier comparison. The simple addition and sigmoid models are likely to have slightly higher stress field errors than the baseline model for a given training session, and the $tan^{-1}$ method is likely to have the same average stress field errors as the baseline model. All PBR models are likely to lower $\text{MSE}_{equil}$ for a given training session, with the simple addition and sigmoid methods by a significant amount.

\begin{table}[h]
\captionsetup{width=0.9\linewidth}
\caption{Average performance and variation from 30 training sessions for each method.}
\centering
\setlength\tabcolsep{10pt}
\begin{tabular}{l c c c}
\hline\hline
&  $\text{MSE}_{\sigma}$ &  Iteration & $\text{MSE}_{equil}$ \\ [0.5ex]
\hline
Baseline & 0.01614 $\pm$0.00050 & 66,000 $\pm$22,000 & 2.348e-4 $\pm$5.94e-5 \\
Sigmoid & 0.01630 $\pm$0.00036 & 70,000 $\pm$22,000 & 6.788e-5 $\pm$1.42e-5 \\
Simple Addition& 0.01663 $\pm$0.00056 & 52,000 $\pm$18,000 & 5.017e-5 $\pm$1.26e-5 \\
$Tan^{-1}$ & 0.01613 $\pm$0.00044 & 59,000 $\pm$19,000 & 1.056e-4 $\pm$8.42e-6 \\
\hline
\end{tabular}
\label{tab:avg_perform}
\end{table}

\begin{table}[h]
\captionsetup{width=0.9\linewidth}
\caption{Average errors shown in Table \ref{tab:avg_perform} normalized to the Baseline error.}
\centering
\setlength\tabcolsep{10pt}
\begin{tabular}{l c c}
\hline\hline
&  $\text{MSE}_{\sigma}$ &  $\text{MSE}_{equil}$ \\ [0.5ex]
\hline
Sigmoid & 1.01  & 0.27 \\
Simple addition& 1.03 & 0.20 \\
$Tan^{-1}$ & 1.00 & 0.43 \\
\hline
\end{tabular}
\label{tab:norm_perf}
\end{table}

\begin{figure}[H]
\centering
\captionsetup{width=.9\linewidth}
\includegraphics[width=0.55\textwidth]{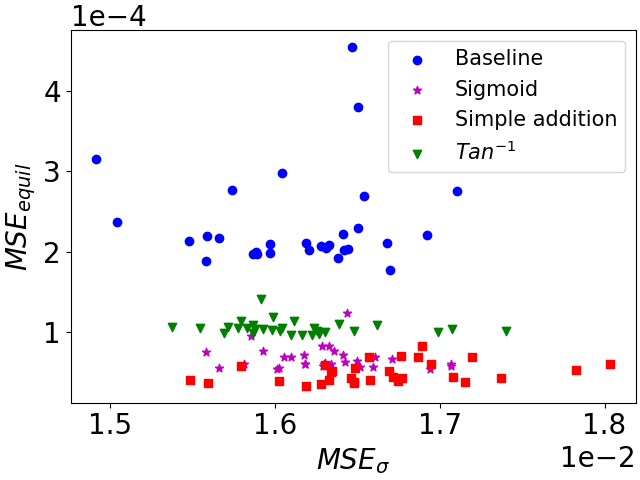}
\caption{The average $\text{MSE}_{\sigma}$ and $\text{MSE}_{equil}$ for all 30 training sessions for each method.}
\label{scatter}
\end{figure}

Figure \ref{scatter} is a scatter plot of the $\text{MSE}_{\sigma}$ vs. $\text{MSE}_{equil}$ from every training session for all methods. This plot further demonstrates the variability in the error metrics for each method. The baseline and simple addition methods have a wider spread of values for $\text{MSE}_{\sigma}$ than the $tan^{-1}$ and sigmoid methods. All PBR methods are very consistent in the $\text{MSE}_{equil}$ across all training sessions.

To better quantify the variability for each method, a bootstrap analysis \cite{efron1994boot} was performed for $\text{MSE}_{\sigma}$, $\text{MSE}_{equil}$, and the convergence iteration. Sampling sizes ranged from [2,30] to estimate the model variation for a given number of training sessions within that range. For a given analysis, the average variation for a metric is calculated from 10,000 samplings with the sample size corresponding to the number of training sessions. All samplings are done with replacement and sample from all 30 training sessions. For example, to estimate the average variation in $\text{MSE}_{\sigma}$ for 3 training sessions, 3 $\text{MSE}_{\sigma}$ values are randomly chosen from all 30 training sessions (with replacement, meaning values can be repeated) to make up one sample. A standard deviation of $\text{MSE}_{\sigma}$ can be calculated for that single sample. This is repeated 9,999 more times to get a total of 10,000 standard deviations from the 10,000 samplings and the average standard deviation can be determined for training a network 3 separate times. To estimate the $\text{MSE}_{\sigma}$ variation of 4 training sessions, a single sample is made up of 4 random samplings with replacement from the 30 training sessions, instead of 3.

\begin{figure}[H]
\centering
\captionsetup{width=\linewidth}
\includegraphics[width=\textwidth]{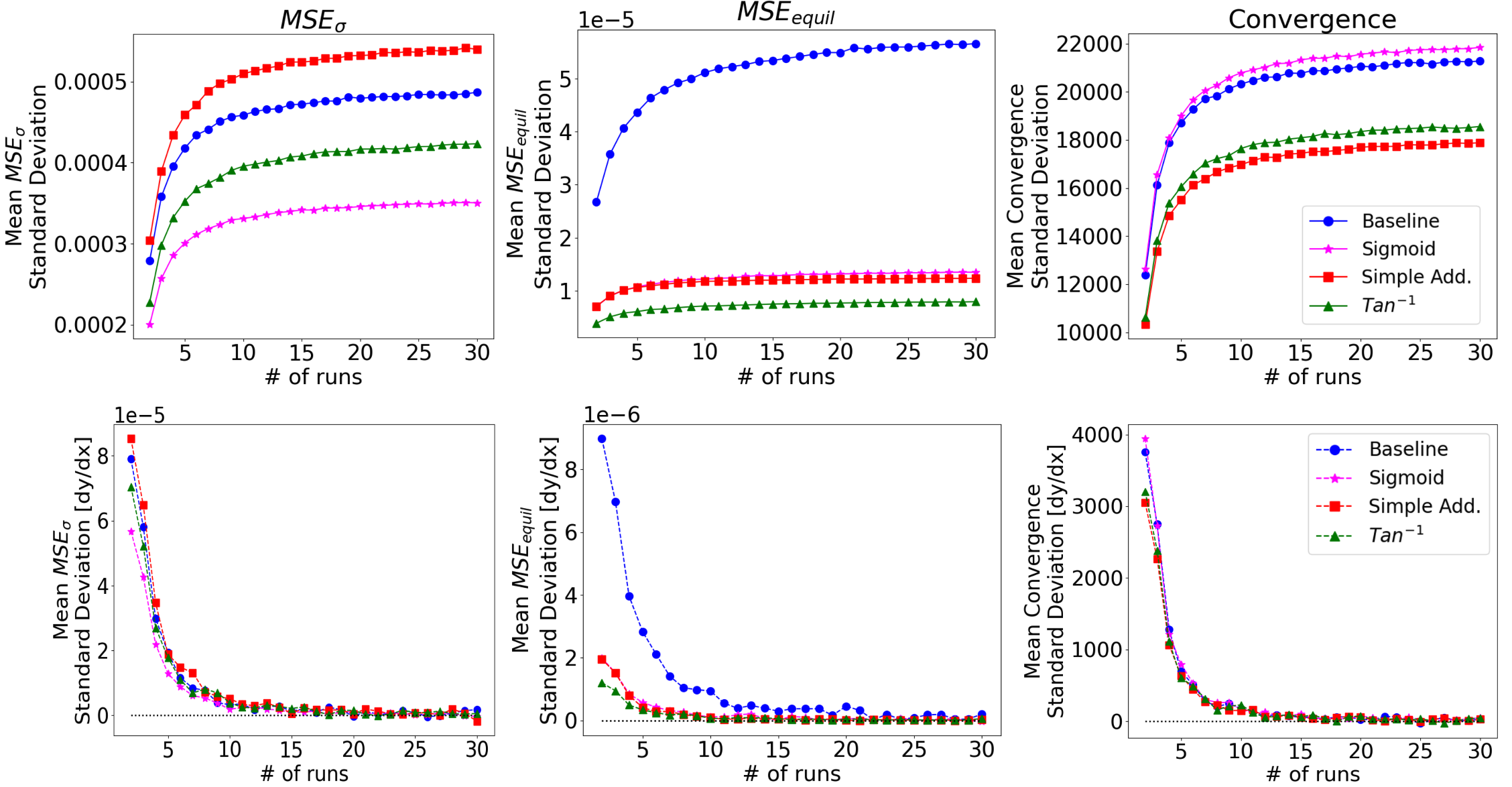}
\caption{Top row: Bootstrap analysis to measure the performance variation of average $\text{MSE}_{\sigma}$, $\text{MSE}_{equil}$, and convergence iteration as a function of number of training sessions. 10,000 samples were taken for each number of training sessions (sample size). Bottom row: the derivative of the curves plotted in the top row (corresponding column-wise). A line at 0 is plotted in the derivative plots to estimate when the derivative converges to 0. Each row corresponds to the legends on the right for their respective row.}
\label{bootstrap}
\end{figure}

The top row of Figure \ref{bootstrap} shows results of the bootstrap estimates of the variation in $\text{MSE}_{\sigma}$, $\text{MSE}_{equil}$, and convergence iteration for 2-30 training sessions. The derivatives are plotted in the bottom row below the corresponding metric. The average standard deviation is converged once the derivative reaches zero. For all metrics, the derivatives for all methods converge to zero at a little over 15 training sessions. This suggests that approximately 15 training sessions are sufficient to measure the variability of a model. All PBR methods significantly reduce the variability in $\text{MSE}_{equil}$. The simple addition method increases the variability in $\text{MSE}_{\sigma}$ compared to the baseline, while the sigmoid and $tan^{-1}$ methods reduce it. The simple addition and $tan^{-1}$ methods reduce the number of iterations needed to reduce $\text{MSE}_{\sigma}$ compared to the baseline, with the sigmoid model slightly increasing the number of iterations needed. 

The average errors from the best, median, and worst performing ``best iteration'' from the 30 training sessions are shown in Table \ref{tab:BMW_perform} for each method. The best iteration is defined as the iteration that obtains the lowest average $\text{MSE}_{\sigma}$ from the test dataset for a given training session. The values shown in Table \ref{tab:BMW_perform} are the prediction error averages and standard deviations of a single training session's best iteration across the test dataset (which is different from the standard deviations in Table \ref{tab:avg_perform}, that measure the deviation in average model performance across different training sessions). Across 30 different training sessions, each method had similar average performances in $\text{MSE}_{\sigma}$, with the simple addition method being the most likely to have the largest stress errors, but the best equilibrium errors. The baseline model resulted in the model with the lowest average $\text{MSE}_{\sigma}$, but significantly higher $\text{MSE}_{equil}$ than the PBR models. The sigmoid and $tan^{-1}$ models found a better balance in optimizing the stress and equilibrium errors with the sigmoid method prioritizing equilibrium errors and $tan^{-1}$ the stress field errors. 
Similar trends were observed in the previous work, although slightly different average performance errors were found. Table \ref{10-30_comp} lists the percent difference from the average performance in $\text{MSE}_{\sigma}$, $\text{MSE}_{equil}$, and convergence across 10 training sessions versus 30 training sessions.


\begin{table}[h]
\captionsetup{width=\linewidth}
\caption{The average $\text{MSE}_{\sigma}$ and $\text{MSE}_{equil}$ from the test dataset for the best, median, and worst performance of the 30 training sessions for each method. The standard deviation of error within the test dataset is shown next to the average errors for that single training session. Note that the number of training sessions is an even number which results in two medians. This table shows the results of the better-performing median. }
\centering
\setlength\tabcolsep{10pt}
\begin{tabular}{l l c c c}
\hline\hline
& &  $\text{MSE}_{\sigma}$ &  Iteration & $\text{MSE}_{equil}$ \\ [0.5ex]
\hline
\multirow{ 3}{*}{Baseline} & Best & 0.01492 $\pm$0.0349 & 24,000 & 3.153e-4 $\pm$5.98e-5 \\
& Median &  0.01621 $\pm$0.0362 & 70,000 & 2.019e-4 $\pm$5.83e-5 \\
& Worst & 0.01711 $\pm$0.0437 & 51,000 & 2.761e-4 $\pm$3.02e-5 \\
\\
\multirow{ 3}{*}{Sigmoid} & Best & 0.01558 $\pm$0.0403 & 59,000 & 7.512e-5 $\pm$2.19e-5 \\
& Median & 0.01630 $\pm$0.0400 & 99,000 & 6.100e-5 $\pm$1.88e-5 \\
& Worst & 0.01707 $\pm$0.0467 & 94,000 & 5.827e-5 $\pm$1.89e-5 \\
\\
\multirow{ 3}{*}{\begin{minipage}[t]{0.1\columnwidth}
    Simple Addition
\end{minipage}} & Best & 0.01548 $\pm$0.0311 & 49,000 & 4.032e-5 $\pm$1.32e-5 \\
& Median & 0.01657 $\pm$0.0363 & 69,000 & 6.949e-5 $\pm$2.03e-5 \\
& Worst & 0.01803 $\pm$0.0429 & 52,000 & 6.072e-5 $\pm$1.81e-5 \\
\\
\multirow{ 3}{*}{$Tan^{-1}$} & Best & 0.01538 $\pm$0.0372 & 88,000 & 1.067e-4 $\pm$3.08e-5 \\
& Median & 0.01603 $\pm$0.0392 & 63,000 & 1.021e-4 $\pm$2.88e-5 \\
& Worst &  0.01740 $\pm$0.0430 & 67,000 & 1.017e-4 $\pm$3.02e-5 \\
\hline
\end{tabular}
\label{tab:BMW_perform}
\end{table}

\begin{table}[h]
\captionsetup{width=\linewidth}
\caption{The difference in average performance for various error metrics and convergence when training each method for an additional 20 training runs.}
\centering
\setlength\tabcolsep{10pt}
\begin{tabular}{l c c c c c}
\hline\hline
& \# of runs &  Baseline & Sigmoid & \makecell{Simple \\ Addition} & $Tan^{-1}$ \\ [0.5ex]
\hline
\multirow{ 3}{*}{$\text{MSE}_{\sigma}$} & 10 & 0.0162 & 0.0163 & 0.0167 & 0.0160\\
& 30 & 0.0161 & 0.0163 & 0.0166 & 0.0161\\
& \% difference & -0.617 & 0.0 & -0.599 & 0.625 \\
\\
\multirow{ 3}{*}{$\text{MSE}_{equil}$} & 10 & 2.19e-4 & 6.62e-5 & 5.36e-5 & 1.07e-4\\
& 30 & 2.35e-4 & 6.79e-5 & 5.02e-5 & 1.06e-4\\
& \% difference & 7.31 & 2.57 & -6.34 & -0.943 \\
\\
\multirow{ 3}{*}{\makecell{Iterations to \\ reduce $\text{MSE}_{\sigma}$}} & 10 & 72,000 & 73,000 & 53,000 & 54,000\\
& 30 & 66,000 & 70,000 & 52,000 & 59,000\\
& \% difference & -8.3 & -4.1 & -1.9 & 9.3 \\
\\
\hline
\end{tabular}
\label{10-30_comp}
\end{table}

\begin{figure}[H]
\centering
\captionsetup{width=.9\linewidth}
\includegraphics[width=0.88\textwidth]{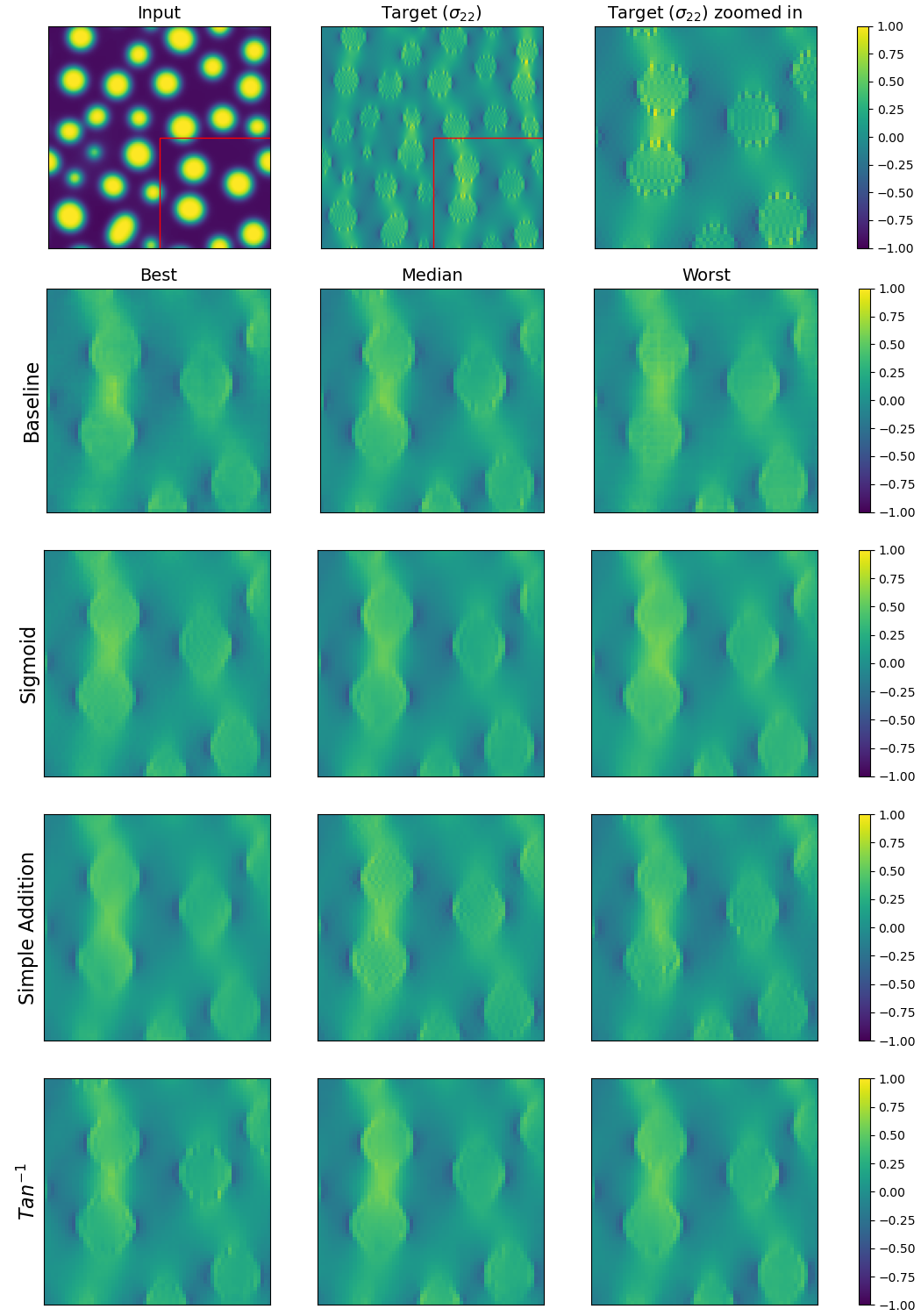}
\caption{The $\sigma_{22}$ stress field (loading direction) for the best, median, and worst performing training session for each method. The stress fields shown are zoomed in on the portion outlined by the red square shown in the top left.}
\label{stress fields}
\end{figure}

Figure \ref{stress fields} shows the generated stress field in the loading direction ($\sigma_{22}$) for each of the models listed in Table \ref{tab:BMW_perform}. This figure shows that for a given training session, the network may capture different features from the dataset. The top-middle image in Figure \ref{stress fields} shows $\sigma_{22}$ from the test dataset, and a zoomed-in portion of it to the right. This stress field exhibits Gibbs oscillations \cite{GibbsPhen} (the pixelated regions around the phase boundaries and going across the yellow circular phase corresponding to the input image), which is an artifact of the CP-FFT simulation. These features are a result of the high elastic contrast in the composite. The baseline method has gridding artifacts in its worst-performing model. In the median performing model, this gridding goes away and instead has a slightly pixelated phase boundary, indicating that it is trying to replicate the Gibbs oscillations. The baseline's best-performing model seems to smooth out the Gibbs oscillations. The worst and median sigmoid models have a slightly more pronounced border at the phase boundary than the best-performing model but seem to otherwise look the same. The simple addition method replicates the Gibbs oscillations in its worst- and median-performing models more prominently than any other method. The simple addition's median-performing model looks the most similar to the oscillations shown in the target but then smooths them out completely in its best-performing model. The $tan^{-1}$ method seems to have the opposite trend, where the Gibbs oscillations become more prominent as $\text{MSE}_{\sigma}$ decreases.  

Note that these Gibbs oscillations are more likely to appear later on in training (though not always), which generally doesn't align with a model's lowest $\text{MSE}_{\sigma}$ for a given training session. In addition, some methods were more likely to replicate the Gibbs oscillations than others. For the same input shown in Figure \ref{stress fields}, each training session was sampled every 9,000 iterations (from the saved checkpoints) to see the prediction progression. An example of this progression is shown in the Appendix for each method. Each training session was visually inspected to evaluate if the Gibbs oscillations appeared at all throughout the iterations sampled. For example, the best performing simple addition model, the best and worst baseline models, and the best, median, and worst sigmoid models would be considered as not having Gibbs oscillations, while the best $tan^{-1}$, the median baseline, and the median and worst simple addition models would be considered as having replicated Gibbs oscillations. For the test example listed in Figure \ref{stress fields}, the $tan^{-1}$ and simple addition methods replicated the Gibbs oscillations at some point during a training session 90\% of the time, the baseline 67\% and sigmoid 57\%.

\begin{figure}[H]
\centering
\captionsetup{width=\linewidth}
\includegraphics[width=0.86\textwidth]{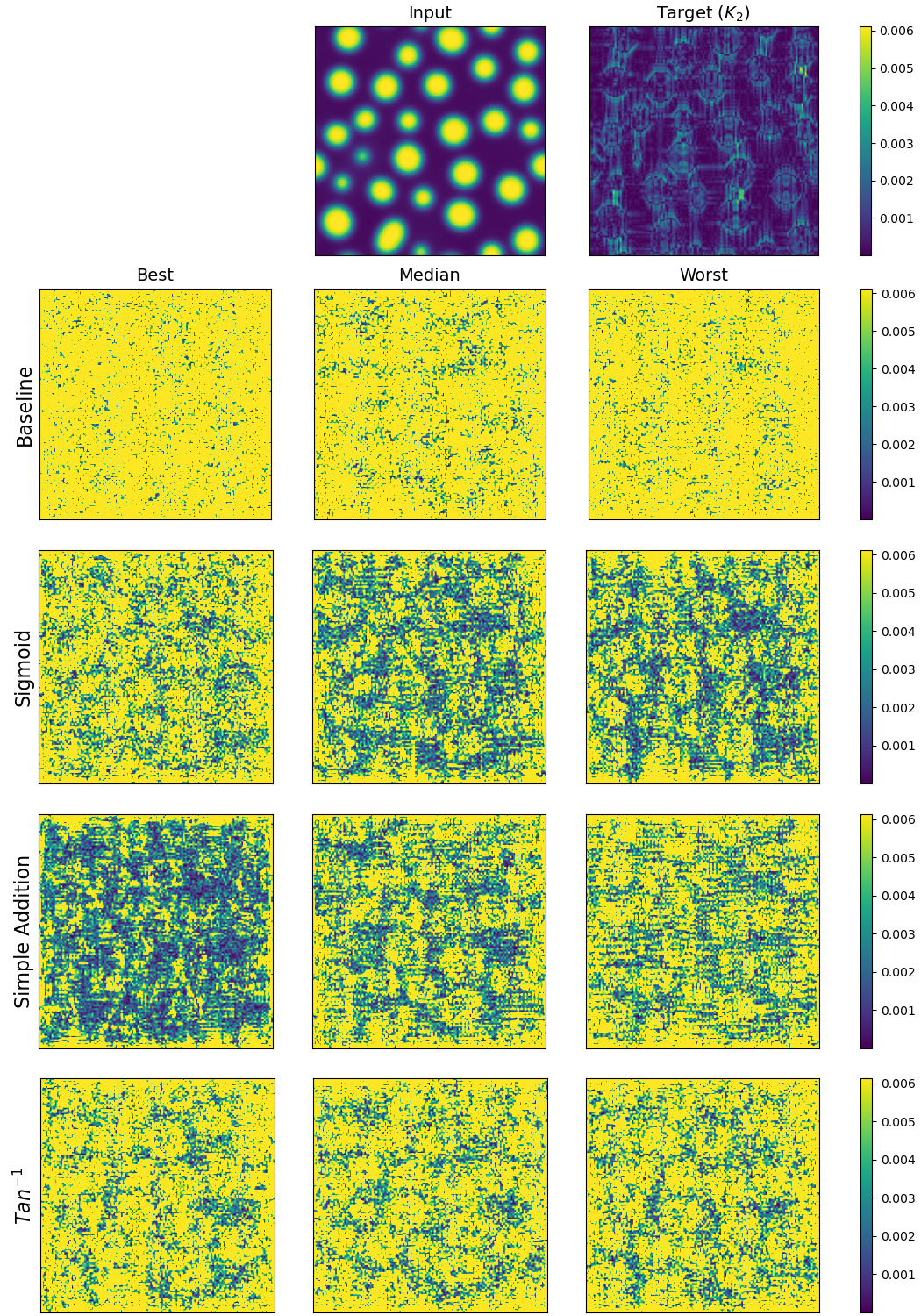}
\caption{The absolute value of a divergence field ($K_2$ from Equation 3 in Ref. \cite{lenau2024importance} is shown here) for the best, median, and worst performing training session for each method. The divergence fields are scaled to the target divergence fields' minimum and maximum values. Yellow pixels indicate a value greater than or equal to the target's largest deviation from equilibrium.}
\label{div fields}
\end{figure}

Figure \ref{div fields} shows the absolute value of a divergence field for the target and the stress fields shown in Figure \ref{stress fields}. The divergence fields calculated from the generated stress fields are scaled to the maximum of the target divergence field. Any yellow pixel in the predicted divergence fields indicates a value greater than or equal to the target's largest deviation from equilibrium. The baseline method has values that are nearly all greater than the target maximum for its best, median, and worst model. The deviations from equilibrium in the predictions seem to be larger than the example shown in previous work, which is likely a result of the target's equilibrium field having larger equilibrium errors as well. The $tan^{-1}$ method has fewer values outside the target range compared to the baseline, but more than the other two PBR models. The $tan^{-1}$ method's divergence fields are consistent across best, median, and worst performing models. The divergence fields for the simple addition method deviate more from the target as the performance gets worse. The opposite trend is observed for the sigmoid method, where the divergence deviates more from the target as the performance improves, i.e. as the $\text{MSE}_{\sigma}$ improves. This may result from a loss competition in the sigmoid method between the stress field error and the equilibrium error. The sigmoid method may improve $\text{MSE}_{\sigma}$ at the expense of $\text{MSE}_{equil}$ for some training sessions.

\section{Discussion}

The goal for developing our network models was to create a more reproducible training strategy that consistently reduces the stress field error ($\text{MSE}_{\sigma}$) and stress equilibrium error ($\text{MSE}_{equil}$) across different training runs.  Thus, we discuss the behavior of the developed networks relative to these goals.

The variation in stress field error, equilibrium error, and convergence across 30 separate training sessions was studied for models with and without PBR. As shown in our previous work, all PBR methods decreased the equilibrium errors for the predicted stress fields when compared to the baseline method. This work shows that it additionally reduces the \textit{variation} of equilibrium error across different training sessions. Note that the simple addition and baseline methods have the greatest variation in $\text{MSE}_{\sigma}$ across training sessions, as shown by the standard deviation values in Table \ref{tab:avg_perform} and Figures \ref{scatter} \& \ref{bootstrap}, but likely for very different reasons. 
The simple addition method may have a larger $\text{MSE}_{\sigma}$ variation since it seems to prioritize $\text{MSE}_{equil}$ which may be over-constraining the network.
The PBR weight may be too strong such that the network sometimes minimizes $\text{MSE}_{equil}$ at the expense of $\text{MSE}_{\sigma}$, resulting in a network that produces stress fields that lower $\text{MSE}_{equil}$ without considering $\text{MSE}_{\sigma}$.  At other times it may learn that reducing $\text{MSE}_{\sigma}$ also reduces $\text{MSE}_{equil}$, which may explain the larger $\text{MSE}_{\sigma}$ variability.
Conversely, the baseline model has nothing to enforce equilibrium within its predictions, resulting in the greatest equilibrium error variation of the evaluated models. 

A key takeaway from this work is that having only a few training sessions is not sufficient for measuring the variability in training a network, as highlighted by the bootstrap analysis. On average, the network needs to be trained a little over 15 times to adequately estimate the variability. After 15 training sessions, the variation appears to converge for all methods and metrics. This estimate also shows the probability of getting a more ``lucky'' run by re-training the same network one more time. When having less than 5 training runs, the probability that the training session having the lowest errors has already been trained is smaller and it may be worthwhile to train the network a few more times to achieve a training session that will result in better performance. However, after about 15 training sessions, finding a run that will have much better (or worse) performance is not likely.

As previously mentioned, Gibbs oscillations are an artifact of the CP-FFT solver used to generate the datasets, and all the training data contains these features. Since the oscillations are an artifact of a simulation, one could argue that it could be advantageous that a network would ignore them. However, the opposing argument is that the Gibbs oscillations appear in the training dataset, therefore they should show up in the network predictions. Furthermore, if a network were to be trained using microstructures having high-frequency features similar to Gibbs oscillations, a representation of those features would be necessary for the network to provide accurate predictions. Regardless, it is shown by Figure \ref{stress fields} that models trained using the same dataset, hyperparameters, and loss function can result in not only different prediction quality but different feature representations, especially concerning the Gibbs oscillations feature. 

Whether a network represents Gibbs oscillations in its predictions depends on the capability of the network to replicate them and when the lowest $\text{MSE}_{\sigma}$ occurs. It was shown that the networks typically reduce the $\text{MSE}_{\sigma}$ before the network learns the Gibbs oscillations. Ref. \cite{Basri2019} showed that neural networks will tend to learn lower frequency features (stress fields that scale with microstructure phase scale) before learning higher frequency features (such as the smaller scale Gibbs oscillations), a trend that is possibly observed for the models studied here. However, there seems to be a difference in how often and how well models replicate the Gibbs oscillations. The simple addition model replicates them very well and often, while the sigmoid model can replicate them well at the end of training (see Supplemental Figures), but not nearly as often. The baseline and $tan^{-1}$ models do not replicate the Gibbs oscillations very well, and the $tan^{-1}$ method does so more often than the sigmoid and baseline models. There is the possibility that the training sessions that did not replicate the Gibbs oscillations just needed more iterations to learn the high-frequency features, as \cite{pham2021} showed that the same network can have different convergence times for different features. A PBR term enforcing similar high-frequency features between targets and predictions could possibly reduce the variability in feature representation between different training sessions; however, that possibility was not investigated.

The Pix2Pix network has shown to be an exemplar for studying the effects of PBR on training variability. However, our findings and general approach are not specific to the Pix2Pix architecture discussed in this work. 
The following is a more general discussion on the variability in training a ML network that is applicable to any ML model. The consequences of a training process having high variability, when to consider the variability, and suggested best practices are discussed, using the above results as an example.
\\
\\
\textbf{What are the implications of high training variability?}\\
Measuring the variability in a model across different training sessions evaluates the reliability of the training method. A higher variability in the average performance of a model means that the reproducibility of the training process is low. Someone using that training process is less likely to achieve similar results to those reported without running the model multiple times and, depending on the model, this could end up being an unreasonable amount of computation time. A repeatable process for training a network is as important as curating a good dataset for training, and the reliability of a training process should be reported to evaluate the reproducibility. The main consequence of a model having high-performance variation is that the performance of a network may not be what is expected, which could lead to misleading conclusions about networks or training algorithms. 
\\
\\
\noindent \textbf{Considerations for choosing the best algorithm/training method}\\
When comparing and choosing the best method (in this case the network loss function), one needs to consider not only which method reduces the errors most efficiently, but how reliably the errors are reduced. One could simply train a method many times and choose the version that reduces the error the most. In this case, the simple addition may be the best pick, since it reduces both $\text{MSE}_{\sigma}$ and $\text{MSE}_{equil}$ (the baseline and $tan^{-1}$ reduce the $\text{MSE}_{\sigma}$ more, but have greater equilibrium errors). However, training a network many times is often not practical depending on how expensive the network is to train. Training a network that takes days, or even months, many times to get the best version is unreasonable. In this case, a low variability network would be more desirable so that training sessions can be minimized to achieve the expected outcome. Low training variability could possibly be advantageous if a network is going to be trained on different datasets for the same reasons, although this should be studied in more depth. The feature representation in network predictions also needs to be considered, as some models may more reliably replicate features than others. 

In summary, choosing the best overall network and training algorithm depends on the computational resources available for training, variability, and accuracy of the network's predictions, and may not be the same network in every scenario. For example, the simple addition may be the best pick if training times are quick and/or sufficient computational resources are available. The simple addition method sufficiently reduces both $\text{MSE}_{\sigma}$ and $\text{MSE}_{equil}$ (the baseline and $tan^{-1}$ reduce the $\text{MSE}_{\sigma}$ more, but have greater equilibrium errors) where the network can be trained many times until a training session is successful. Alternatively, if computing resources are limited, the sigmoid method may be the best option. The sigmoid method has the lowest $\text{MSE}_{\sigma}$ among the worst-performing models and is consistent in reproducing similar $\text{MSE}_{equil}$ values.
\\
\\
\noindent \textbf{Suggested practices}\\
Reporting the average performance variation of a network becomes especially important when comparing deep learning methods. 
A large motivation behind incorporating physics into a deep learning network is to train the network more efficiently and/or result in a more accurate network. Unless the variability is reported, the comparative improvement in accuracy/efficiency becomes obscured. This work clearly demonstrates that comparing just one or a small number of training runs for a method may result in misleading conclusions about the performance of a network. For example, comparing only the best performances of the sigmoid and $tan^{-1}$ methods would lead to the conclusion that the $tan^{-1}$ method converges more slowly than the sigmoid method but achieves a lower error, when the opposite is true on average (see Figure \ref{bootstrap} and Table \ref{tab:BMW_perform}). 
Comparing the baseline and $tan^{-1}$ methods, both have similar average performance in stress errors, yet the baseline method's best-performing model has a lower $\text{MSE}_{\sigma}$ than the $tan^{-1}$ method's best-performing model. A similar situation is seen comparing the best-performing models for the simple addition and sigmoid methods. The simple addition method's best-performing model achieves lower errors than the sigmoid method's best, but the sigmoid method will on average achieve a lower $\text{MSE}_{\sigma}$. 

A bootstrap analysis was useful in estimating the variation of each metric for each network, as well as determining the number of training sessions needed to estimate these variations. When computing resources allow, the authors suggest performing this analysis to ensure the variation values are converged. However, we acknowledge that with limited computing resources, or networks requiring very long training times, the number of training sessions to converge the variation measurement may not be reasonable. This poses an issue for fair comparisons of networks with expensive training times.

Arguably, reporting the variability of a model may become less important when a comparison among deep learning methods, parameters, or loss functions is \textit{not} made, and code and/or a trained model is provided. Although a measure of the variability would still be helpful for someone else wanting to repeat the training process, measuring the variability of a network might be difficult to justify if computing resources are limited. For the most reproducible model, providing code and hyper-parameters used for training is the most straightforward method.

\section{Conclusions}
The model variation was studied across different training sessions of networks having, and without, physics-informed losses to enforce stress equilibrium. All physics-informed losses reduced the variation in equilibrium error compared to the baseline, and two of the three physics-informed losses also reduced the variation in stress errors. This shows that a physics-informed loss can create a more reliable and repeatable network. Two of the three physics-informed losses reduced the convergence and variation of convergence iteration of the stress error compared to the baseline. Also shown is that networks will vary with regard to which features are captured for a given training session and that some methods were more likely to reproduce high-frequency features than others. Therefore, reporting model/network variability is important for comparing models.  Our understanding is that the aspects of model variability during training, and the resultant influences on the outcome of a fully trained model as observed in the present study, are likely applicable any time that deep learning tools are developed for the relatively small datasets that are common in the materials sciences.  Ideally, these aspects should be reported along with model results.

\backmatter

\bmhead{Acknowledgements}

This study was funded by National Science Foundation award CMMI 1826149 (Quantitative Representation of Microstructure). AL and SRN also received funding from the National Science Foundation award EEC 2133630 (NSF Engineering Research Center for Hybrid Autonomous Manufacturing Moving from Evolution to Revolution (ERC-HAMMER)). The funder played no role in study design, data collection, analysis and interpretation of data, or the writing of this manuscript. This work was performed, in part, at the Center for Integrated Nanotechnologies, an Office of Science User Facility operated for the U.S. Department of Energy (DOE) Office of Science by Los Alamos National Laboratory (Contract 89233218CNA000001) and Sandia National Laboratories (Contract DE-NA-0003525). 

\bmhead{Author Contributions}
AL developed and implemented the ML Network and performed the numerical experiments. All authors contributed to the initial experimental design and subsequent iterations. All authors contributed to discussions and interpretation of the results. AL composed the early drafts of the manuscript. SRN and DMD provided revision and feedback on early draft and final revisions on submitted version. All authors reviewed the final manuscript and approve its submission. 

\bmhead{Competing Interests}
All authors declare no financial or non-financial competing interests. 

\bmhead{Code Availability}
The underlying code for this study is available in the PB-GAN repository and can be accessed via this link https://github.com/mesoOSU/PB-GAN. All training, validation, and test datasets necessary to reproduce the current study are also available through the PB-GAN repository.

\bmhead{Ethics approval and consent to participate}
Not applicable.
\bmhead{Consent for publication}
Not applicable.
\bmhead{Materials availability}
Not applicable.
\bmhead{Data availability}
Not applicable (with code availability).



\bibliography{newest_manuscript}


\begin{thebibliography}{35}
\ifx \bisbn   \undefined \def \bisbn  #1{ISBN #1}\fi
\ifx \binits  \undefined \def \binits#1{#1}\fi
\ifx \bauthor  \undefined \def \bauthor#1{#1}\fi
\ifx \batitle  \undefined \def \batitle#1{#1}\fi
\ifx \bjtitle  \undefined \def \bjtitle#1{#1}\fi
\ifx \bvolume  \undefined \def \bvolume#1{\textbf{#1}}\fi
\ifx \byear  \undefined \def \byear#1{#1}\fi
\ifx \bissue  \undefined \def \bissue#1{#1}\fi
\ifx \bfpage  \undefined \def \bfpage#1{#1}\fi
\ifx \blpage  \undefined \def \blpage #1{#1}\fi
\ifx \burl  \undefined \def \burl#1{\textsf{#1}}\fi
\ifx \doiurl  \undefined \def \doiurl#1{\url{https://doi.org/#1}}\fi
\ifx \betal  \undefined \def \betal{\textit{et al.}}\fi
\ifx \binstitute  \undefined \def \binstitute#1{#1}\fi
\ifx \binstitutionaled  \undefined \def \binstitutionaled#1{#1}\fi
\ifx \bctitle  \undefined \def \bctitle#1{#1}\fi
\ifx \beditor  \undefined \def \beditor#1{#1}\fi
\ifx \bpublisher  \undefined \def \bpublisher#1{#1}\fi
\ifx \bbtitle  \undefined \def \bbtitle#1{#1}\fi
\ifx \bedition  \undefined \def \bedition#1{#1}\fi
\ifx \bseriesno  \undefined \def \bseriesno#1{#1}\fi
\ifx \blocation  \undefined \def \blocation#1{#1}\fi
\ifx \bsertitle  \undefined \def \bsertitle#1{#1}\fi
\ifx \bsnm \undefined \def \bsnm#1{#1}\fi
\ifx \bsuffix \undefined \def \bsuffix#1{#1}\fi
\ifx \bparticle \undefined \def \bparticle#1{#1}\fi
\ifx \barticle \undefined \def \barticle#1{#1}\fi
\bibcommenthead
\ifx \bconfdate \undefined \def \bconfdate #1{#1}\fi
\ifx \botherref \undefined \def \botherref #1{#1}\fi
\ifx \url \undefined \def \url#1{\textsf{#1}}\fi
\ifx \bchapter \undefined \def \bchapter#1{#1}\fi
\ifx \bbook \undefined \def \bbook#1{#1}\fi
\ifx \bcomment \undefined \def \bcomment#1{#1}\fi
\ifx \oauthor \undefined \def \oauthor#1{#1}\fi
\ifx \citeauthoryear \undefined \def \citeauthoryear#1{#1}\fi
\ifx \endbibitem  \undefined \def \endbibitem {}\fi
\ifx \bconflocation  \undefined \def \bconflocation#1{#1}\fi
\ifx \arxivurl  \undefined \def \arxivurl#1{\textsf{#1}}\fi
\csname PreBibitemsHook\endcsname

\bibitem[\protect\citeauthoryear{Ge et~al.}{2020}]{DL_micro}
\begin{barticle}
\bauthor{\bsnm{Ge}, \binits{M.}},
\bauthor{\bsnm{Su}, \binits{F.}},
\bauthor{\bsnm{Zhao}, \binits{Z.}},
\bauthor{\bsnm{Su}, \binits{D.}}:
\batitle{Deep learning analysis on microscopic imaging in materials science}.
\bjtitle{Materials Today Nano}
\bvolume{11},
\bfpage{100087}
(\byear{2020})
\doiurl{10.1016/j.mtnano.2020.100087}
\end{barticle}
\endbibitem

\bibitem[\protect\citeauthoryear{Jacobs}{2022}]{JACOBS2022}
\begin{barticle}
\bauthor{\bsnm{Jacobs}, \binits{R.}}:
\batitle{Deep learning object detection in materials science: Current state and future directions}.
\bjtitle{Computational Materials Science}
\bvolume{211},
\bfpage{111527}
(\byear{2022})
\doiurl{10.1016/j.commatsci.2022.111527}
\end{barticle}
\endbibitem

\bibitem[\protect\citeauthoryear{Merchant et~al.}{2023}]{Merchant2023}
\begin{barticle}
\bauthor{\bsnm{Merchant}, \binits{A.}},
\bauthor{\bsnm{Batzner}, \binits{S.}},
\bauthor{\bsnm{Schoenholz}, \binits{S.S.}},
\bauthor{\bsnm{Aykol}, \binits{M.}},
\bauthor{\bsnm{Cheon}, \binits{G.}},
\bauthor{\bsnm{Cubuk}, \binits{E.D.}}:
\batitle{Scaling deep learning for materials discovery}.
\bjtitle{Nature}
\bvolume{624},
\bfpage{80}--\blpage{85}
(\byear{2023})
\end{barticle}
\endbibitem

\bibitem[\protect\citeauthoryear{Zuo et~al.}{2021}]{ZUO2021}
\begin{barticle}
\bauthor{\bsnm{Zuo}, \binits{Y.}},
\bauthor{\bsnm{Qin}, \binits{M.}},
\bauthor{\bsnm{Chen}, \binits{C.}},
\bauthor{\bsnm{Ye}, \binits{W.}},
\bauthor{\bsnm{Li}, \binits{X.}},
\bauthor{\bsnm{Luo}, \binits{J.}},
\bauthor{\bsnm{Ong}, \binits{S.P.}}:
\batitle{Accelerating materials discovery with bayesian optimization and graph deep learning}.
\bjtitle{Materials Today}
\bvolume{51},
\bfpage{126}--\blpage{135}
(\byear{2021})
\doiurl{10.1016/j.mattod.2021.08.012}
\end{barticle}
\endbibitem

\bibitem[\protect\citeauthoryear{Li et~al.}{2019}]{LI2019}
\begin{barticle}
\bauthor{\bsnm{Li}, \binits{X.}},
\bauthor{\bsnm{Liu}, \binits{Z.}},
\bauthor{\bsnm{Cui}, \binits{S.}},
\bauthor{\bsnm{Luo}, \binits{C.}},
\bauthor{\bsnm{Li}, \binits{C.}},
\bauthor{\bsnm{Zhuang}, \binits{Z.}}:
\batitle{Predicting the effective mechanical property of heterogeneous materials by image based modeling and deep learning}.
\bjtitle{Computer Methods in Applied Mechanics and Engineering}
\bvolume{347},
\bfpage{735}--\blpage{753}
(\byear{2019})
\doiurl{10.1016/j.cma.2019.01.005}
\end{barticle}
\endbibitem

\bibitem[\protect\citeauthoryear{Jha et~al.}{2022}]{jha2022}
\begin{barticle}
\bauthor{\bsnm{Jha}, \binits{D.}},
\bauthor{\bsnm{Gupta}, \binits{V.}},
\bauthor{\bsnm{Liao}, \binits{W.-K.}},
\bauthor{\bsnm{Choudhary}, \binits{A.}},
\bauthor{\bsnm{Agrawal}, \binits{A.}}:
\batitle{Moving closer to experimental level materials property prediction using ai}.
\bjtitle{Sci Rep}
\bvolume{12}(\bissue{3}),
\bfpage{11953}
(\byear{2022})
\doiurl{10.1038/s41598-022-15816-0}
\end{barticle}
\endbibitem

\bibitem[\protect\citeauthoryear{Henkes and Wessels}{2022}]{HENKES2022}
\begin{barticle}
\bauthor{\bsnm{Henkes}, \binits{A.}},
\bauthor{\bsnm{Wessels}, \binits{H.}}:
\batitle{Three-dimensional microstructure generation using generative adversarial neural networks in the context of continuum micromechanics}.
\bjtitle{Computer Methods in Applied Mechanics and Engineering}
\bvolume{400},
\bfpage{115497}
(\byear{2022})
\doiurl{10.1016/j.cma.2022.115497}
\end{barticle}
\endbibitem

\bibitem[\protect\citeauthoryear{Chun et~al.}{2020}]{chun2020}
\begin{barticle}
\bauthor{\bsnm{Chun}, \binits{S.}},
\bauthor{\bsnm{Roy}, \binits{S.}},
\bauthor{\bsnm{Nguyen}, \binits{Y.T.}},
\bauthor{\bsnm{Choi}, \binits{J.B.}},
\bauthor{\bsnm{Udaykumar}, \binits{H.S.}},
\bauthor{\bsnm{Baek}, \binits{S.S.}}:
\batitle{Deep learning for synthetic microstructure generation in a materials-by-design framework for heterogeneous energetic materials}.
\bjtitle{Sci Rep}
\bvolume{10},
\bfpage{13307}
(\byear{2020})
\doiurl{10.1038/s41598-020-70149-0}
\end{barticle}
\endbibitem

\bibitem[\protect\citeauthoryear{Agrawal and Choudhary}{2019}]{Agrawal_Choudhary_2019}
\begin{barticle}
\bauthor{\bsnm{Agrawal}, \binits{A.}},
\bauthor{\bsnm{Choudhary}, \binits{A.}}:
\batitle{Deep materials informatics: Applications of deep learning in materials science}.
\bjtitle{MRS Communications}
\bvolume{9}(\bissue{3}),
\bfpage{779}--\blpage{792}
(\byear{2019})
\doiurl{10.1557/mrc.2019.73}
\end{barticle}
\endbibitem

\bibitem[\protect\citeauthoryear{Pham et~al.}{2021}]{pham2021}
\begin{bchapter}
\bauthor{\bsnm{Pham}, \binits{H.V.}},
\bauthor{\bsnm{Qian}, \binits{S.}},
\bauthor{\bsnm{Wang}, \binits{J.}},
\bauthor{\bsnm{Lutellier}, \binits{T.}},
\bauthor{\bsnm{Rosenthal}, \binits{J.}},
\bauthor{\bsnm{Tan}, \binits{L.}},
\bauthor{\bsnm{Yu}, \binits{Y.}},
\bauthor{\bsnm{Nagappan}, \binits{N.}}:
\bctitle{Problems and opportunities in training deep learning software systems: an analysis of variance}.
In: \bbtitle{Proceedings of the 35th IEEE/ACM International Conference on Automated Software Engineering}.
\bsertitle{ASE '20},
pp. \bfpage{771}--\blpage{783}.
\bpublisher{Association for Computing Machinery},
\blocation{New York, NY, USA}
(\byear{2021}).
\doiurl{10.1145/3324884.3416545} .
\burl{https://doi.org/10.1145/3324884.3416545}
\end{bchapter}
\endbibitem

\bibitem[\protect\citeauthoryear{Khan et~al.}{2018}]{khan2019}
\begin{bchapter}
\bauthor{\bsnm{Khan}, \binits{M.M.R.}},
\bauthor{\bsnm{Arif}, \binits{R.B.}},
\bauthor{\bsnm{Siddique}, \binits{M.A.B.}},
\bauthor{\bsnm{Oishe}, \binits{M.R.}}:
\bctitle{Study and observation of the variation of accuracies of knn, svm, lmnn, enn algorithms on eleven different datasets from uci machine learning repository}.
In: \bbtitle{2018 4th International Conference on Electrical Engineering and Information \& Communication Technology (iCEEiCT)},
pp. \bfpage{124}--\blpage{129}
(\byear{2018}).
\doiurl{10.1109/CEEICT.2018.8628041}
\end{bchapter}
\endbibitem

\bibitem[\protect\citeauthoryear{Bouthillier et~al.}{2021}]{bouthillier2021}
\begin{bchapter}
\bauthor{\bsnm{Bouthillier}, \binits{X.}},
\bauthor{\bsnm{Delaunay}, \binits{P.}},
\bauthor{\bsnm{Bronzi}, \binits{M.}},
\bauthor{\bsnm{Trofimov}, \binits{A.}},
\bauthor{\bsnm{Nichyporuk}, \binits{B.}},
\bauthor{\bsnm{Szeto}, \binits{J.}},
\bauthor{\bsnm{Mohammadi~Sepahvand}, \binits{N.}},
\bauthor{\bsnm{Raff}, \binits{E.}},
\bauthor{\bsnm{Madan}, \binits{K.}},
\bauthor{\bsnm{Voleti}, \binits{V.}},
\bauthor{\bsnm{Ebrahimi~Kahou}, \binits{S.}},
\bauthor{\bsnm{Michalski}, \binits{V.}},
\bauthor{\bsnm{Arbel}, \binits{T.}},
\bauthor{\bsnm{Pal}, \binits{C.}},
\bauthor{\bsnm{Varoquaux}, \binits{G.}},
\bauthor{\bsnm{Vincent}, \binits{P.}}:
\bctitle{Accounting for variance in machine learning benchmarks}.
In: \beditor{\bsnm{Smola}, \binits{A.}},
\beditor{\bsnm{Dimakis}, \binits{A.}},
\beditor{\bsnm{Stoica}, \binits{I.}} (eds.)
\bbtitle{Proceedings of Machine Learning and Systems},
vol. \bseriesno{3},
pp. \bfpage{747}--\blpage{769}
(\byear{2021}).
\burl{https://proceedings.mlsys.org/paper_files/paper/2021/file/0184b0cd3cfb185989f858a1d9f5c1eb-Paper.pdf}
\end{bchapter}
\endbibitem

\bibitem[\protect\citeauthoryear{Alahmari et~al.}{2020}]{alahmari2020}
\begin{barticle}
\bauthor{\bsnm{Alahmari}, \binits{S.S.}},
\bauthor{\bsnm{Goldgof}, \binits{D.B.}},
\bauthor{\bsnm{Mouton}, \binits{P.R.}},
\bauthor{\bsnm{Hall}, \binits{L.O.}}:
\batitle{Challenges for the repeatability of deep learning models}.
\bjtitle{IEEE Access}
\bvolume{8},
\bfpage{211860}--\blpage{211868}
(\byear{2020})
\doiurl{10.1109/ACCESS.2020.3039833}
\end{barticle}
\endbibitem

\bibitem[\protect\citeauthoryear{Qian et~al.}{2021}]{qian2021}
\begin{bchapter}
\bauthor{\bsnm{Qian}, \binits{S.}},
\bauthor{\bsnm{Pham}, \binits{V.H.}},
\bauthor{\bsnm{Lutellier}, \binits{T.}},
\bauthor{\bsnm{Hu}, \binits{Z.}},
\bauthor{\bsnm{Kim}, \binits{J.}},
\bauthor{\bsnm{Tan}, \binits{L.}},
\bauthor{\bsnm{Yu}, \binits{Y.}},
\bauthor{\bsnm{Chen}, \binits{J.}},
\bauthor{\bsnm{Shah}, \binits{S.}}:
\bctitle{Are my deep learning systems fair? an empirical study of fixed-seed training}.
In: \beditor{\bsnm{Ranzato}, \binits{M.}},
\beditor{\bsnm{Beygelzimer}, \binits{A.}},
\beditor{\bsnm{Dauphin}, \binits{Y.}},
\beditor{\bsnm{Liang}, \binits{P.S.}},
\beditor{\bsnm{Vaughan}, \binits{J.W.}} (eds.)
\bbtitle{Advances in Neural Information Processing Systems},
vol. \bseriesno{34},
pp. \bfpage{30211}--\blpage{30227}
(\byear{2021}).
\burl{https://proceedings.neurips.cc/paper_files/paper/2021/file/fdda6e957f1e5ee2f3b311fe4f145ae1-Paper.pdf}
\end{bchapter}
\endbibitem

\bibitem[\protect\citeauthoryear{Pinto et~al.}{2022}]{Pinto_Alguacil_Bauerheim_2022}
\begin{barticle}
\bauthor{\bsnm{Pinto}, \binits{W.G.}},
\bauthor{\bsnm{Alguacil}, \binits{A.}},
\bauthor{\bsnm{Bauerheim}, \binits{M.}}:
\batitle{On the reproducibility of fully convolutional neural networks for modeling time–space-evolving physical systems}.
\bjtitle{Data-Centric Engineering}
\bvolume{3},
\bfpage{19}
(\byear{2022})
\doiurl{10.1017/dce.2022.18}
\end{barticle}
\endbibitem

\bibitem[\protect\citeauthoryear{Pineau et~al.}{2021}]{Pineau}
\begin{botherref}
\oauthor{\bsnm{Pineau}, \binits{J.}},
\oauthor{\bsnm{Vincent-Lamarre}, \binits{P.}},
\oauthor{\bsnm{Sinha}, \binits{K.}},
\oauthor{\bsnm{Larivi\`{e}re}, \binits{V.}},
\oauthor{\bsnm{Beygelzimer}, \binits{A.}},
\oauthor{\bsnm{d'Alch\'{e}-Buc}, \binits{F.}},
\oauthor{\bsnm{Fox}, \binits{E.}},
\oauthor{\bsnm{Larochelle}, \binits{H.}}:
Improving reproducibility in machine learning research (a report from the neurips 2019 reproducibility program).
J. Mach. Learn. Res.
\textbf{22}(1)
(2021)
\end{botherref}
\endbibitem

\bibitem[\protect\citeauthoryear{Wang et~al.}{2020}]{best_prac_sparks}
\begin{barticle}
\bauthor{\bsnm{Wang}, \binits{A.Y.-T.}},
\bauthor{\bsnm{Murdock}, \binits{R.J.}},
\bauthor{\bsnm{Kauwe}, \binits{S.K.}},
\bauthor{\bsnm{Oliynyk}, \binits{A.O.}},
\bauthor{\bsnm{Gurlo}, \binits{A.}},
\bauthor{\bsnm{Brgoch}, \binits{J.}},
\bauthor{\bsnm{Persson}, \binits{K.A.}},
\bauthor{},
\bauthor{\bsnm{Sparks}, \binits{T.D.}}:
\batitle{Machine learning for materials scientists: An introductory guide toward best practices}.
\bjtitle{Chemistry of Materials}
\bvolume{32}(\bissue{12}),
\bfpage{4954}--\blpage{4965}
(\byear{2020})
\doiurl{10.1021/acs.chemmater.0c01907}
\end{barticle}
\endbibitem

\bibitem[\protect\citeauthoryear{Shin et~al.}{2023}]{shin_dmn}
\begin{botherref}
\oauthor{\bsnm{Shin}, \binits{D.}},
\oauthor{\bsnm{Alberdi}, \binits{R.}},
\oauthor{\bsnm{Lebensohn}, \binits{R.A.}},
\oauthor{\bsnm{Dingreville}, \binits{R.}}:
Deep material network via a quilting strategy: visualization for explainability and recursive training for improved accuracy.
npj Comput Mater
\textbf{9}(128)
(2023)
\doiurl{10.1038/s41524-023-01085-6}
\end{botherref}
\endbibitem

\bibitem[\protect\citeauthoryear{Tavazza et~al.}{2021}]{tavazza2021}
\begin{barticle}
\bauthor{\bsnm{Tavazza}, \binits{F.}},
\bauthor{\bsnm{DeCost}, \binits{B.}},
\bauthor{\bsnm{Choudhary}, \binits{K.}}:
\batitle{Uncertainty prediction for machine learning models of material properties}.
\bjtitle{ACS Omega}
\bvolume{6}(\bissue{48}),
\bfpage{32431}--\blpage{32440}
(\byear{2021})
\doiurl{10.1021/acsomega.1c03752}
{\href{https://arxiv.org/abs/https://doi.org/10.1021/acsomega.1c03752}{{https://doi.org/10.1021/acsomega.1c03752}}}
\end{barticle}
\endbibitem

\bibitem[\protect\citeauthoryear{Olivier et~al.}{2021}]{OLIVIER2021}
\begin{barticle}
\bauthor{\bsnm{Olivier}, \binits{A.}},
\bauthor{\bsnm{Shields}, \binits{M.D.}},
\bauthor{\bsnm{Graham-Brady}, \binits{L.}}:
\batitle{Bayesian neural networks for uncertainty quantification in data-driven materials modeling}.
\bjtitle{Computer Methods in Applied Mechanics and Engineering}
\bvolume{386},
\bfpage{114079}
(\byear{2021})
\doiurl{10.1016/j.cma.2021.114079}
\end{barticle}
\endbibitem

\bibitem[\protect\citeauthoryear{Viana and Subramaniyan}{2021}]{Viana2021}
\begin{barticle}
\bauthor{\bsnm{Viana}, \binits{F.A.C.}},
\bauthor{\bsnm{Subramaniyan}, \binits{A.K.}}:
\batitle{A survey of bayesian calibration and physics-informed neural networks in scientific modeling}.
\bjtitle{Archives of Computational Methods in Engineering}
\bvolume{28},
\bfpage{3801}--\blpage{3830}
(\byear{2021})
\end{barticle}
\endbibitem

\bibitem[\protect\citeauthoryear{Li et~al.}{2024}]{LI2024}
\begin{barticle}
\bauthor{\bsnm{Li}, \binits{J.}},
\bauthor{\bsnm{Long}, \binits{X.}},
\bauthor{\bsnm{Deng}, \binits{X.}},
\bauthor{\bsnm{Jiang}, \binits{W.}},
\bauthor{\bsnm{Zhou}, \binits{K.}},
\bauthor{\bsnm{Jiang}, \binits{C.}},
\bauthor{\bsnm{Zhang}, \binits{X.}}:
\batitle{A principled distance-aware uncertainty quantification approach for enhancing the reliability of physics-informed neural network}.
\bjtitle{Reliability Engineering \& System Safety}
\bvolume{245},
\bfpage{109963}
(\byear{2024})
\doiurl{10.1016/j.ress.2024.109963}
\end{barticle}
\endbibitem

\bibitem[\protect\citeauthoryear{Lemay et~al.}{2022}]{lemay2022improving}
\begin{botherref}
\oauthor{\bsnm{Lemay}, \binits{A.}},
\oauthor{\bsnm{Hoebel}, \binits{K.}},
\oauthor{\bsnm{Bridge}, \binits{C.P.}},
\oauthor{\bsnm{Befano}, \binits{B.}},
\oauthor{\bsnm{Sanjosé}, \binits{S.D.}},
\oauthor{\bsnm{Egemen}, \binits{D.}},
\oauthor{\bsnm{Rodriguez}, \binits{A.C.}},
\oauthor{\bsnm{Schiffman}, \binits{M.}},
\oauthor{\bsnm{Campbell}, \binits{J.P.}},
\oauthor{\bsnm{Kalpathy-Cramer}, \binits{J.}}:
Improving the repeatability of deep learning models with Monte Carlo dropout
(2022)
\end{botherref}
\endbibitem

\bibitem[\protect\citeauthoryear{Bai and Song}{2023}]{BAI2023}
\begin{barticle}
\bauthor{\bsnm{Bai}, \binits{Z.}},
\bauthor{\bsnm{Song}, \binits{S.}}:
\batitle{Structural reliability analysis based on neural networks with physics-informed training samples}.
\bjtitle{Engineering Applications of Artificial Intelligence}
\bvolume{126},
\bfpage{107157}
(\byear{2023})
\doiurl{10.1016/j.engappai.2023.107157}
\end{barticle}
\endbibitem

\bibitem[\protect\citeauthoryear{Lu et~al.}{2022}]{LU2022}
\begin{barticle}
\bauthor{\bsnm{Lu}, \binits{Y.}},
\bauthor{\bsnm{Wang}, \binits{B.}},
\bauthor{\bsnm{Zhao}, \binits{Y.}},
\bauthor{\bsnm{Yang}, \binits{X.}},
\bauthor{\bsnm{Li}, \binits{L.}},
\bauthor{\bsnm{Dong}, \binits{M.}},
\bauthor{\bsnm{Lv}, \binits{Q.}},
\bauthor{\bsnm{Zhou}, \binits{F.}},
\bauthor{\bsnm{Gu}, \binits{N.}},
\bauthor{\bsnm{Shang}, \binits{L.}}:
\batitle{Physics-informed surrogate modeling for hydro-fracture geometry prediction based on deep learning}.
\bjtitle{Energy}
\bvolume{253},
\bfpage{124139}
(\byear{2022})
\doiurl{10.1016/j.energy.2022.124139}
\end{barticle}
\endbibitem

\bibitem[\protect\citeauthoryear{Krishnapriyan et~al.}{2021}]{Krishnapriyan2021}
\begin{bchapter}
\bauthor{\bsnm{Krishnapriyan}, \binits{A.}},
\bauthor{\bsnm{Gholami}, \binits{A.}},
\bauthor{\bsnm{Zhe}, \binits{S.}},
\bauthor{\bsnm{Kirby}, \binits{R.}},
\bauthor{\bsnm{Mahoney}, \binits{M.W.}}:
\bctitle{Characterizing possible failure modes in physics-informed neural networks}.
In: \beditor{\bsnm{Ranzato}, \binits{M.}},
\beditor{\bsnm{Beygelzimer}, \binits{A.}},
\beditor{\bsnm{Dauphin}, \binits{Y.}},
\beditor{\bsnm{Liang}, \binits{P.S.}},
\beditor{\bsnm{Vaughan}, \binits{J.W.}} (eds.)
\bbtitle{Advances in Neural Information Processing Systems},
vol. \bseriesno{34},
pp. \bfpage{26548}--\blpage{26560}.
\bpublisher{Curran Associates, Inc.}, \blocation{???}
(\byear{2021}).
\burl{https://proceedings.neurips.cc/paper_files/paper/2021/file/df438e5206f31600e6ae4af72f2725f1-Paper.pdf}
\end{bchapter}
\endbibitem

\bibitem[\protect\citeauthoryear{Meng et~al.}{2025}]{meng_rev_2025}
\begin{botherref}
\oauthor{\bsnm{Meng}, \binits{C.}},
\oauthor{\bsnm{Griesemer}, \binits{S.}},
\oauthor{\bsnm{Cao}, \binits{D.}},
\oauthor{\bsnm{Seo}, \binits{S.}},
\oauthor{\bsnm{Liu}, \binits{Y.}}:
When physics meets machine learning: a survey of physics-informed machine learning.
Mach. Learn. Comput. Sci. Eng
\textbf{1}(20)
(2025)
\end{botherref}
\endbibitem

\bibitem[\protect\citeauthoryear{Cuomo et~al.}{2022}]{cuomo_rev2022}
\begin{botherref}
\oauthor{\bsnm{Cuomo}, \binits{S.}},
\oauthor{\bsnm{Cola}, \binits{V.S.D.}},
\oauthor{\bsnm{Giampaolo}, \binits{F.}},
\oauthor{\bsnm{Rozza}, \binits{G.}},
\oauthor{\bsnm{Raissi}, \binits{M.}},
\oauthor{\bsnm{Piccialli}, \binits{F.}}:
Scientific machine learning through physics–informed neural networks: Where we are and what’s next.
J Sci Comput
\textbf{92}(88)
(2022)
\end{botherref}
\endbibitem

\bibitem[\protect\citeauthoryear{Blechschmidt and Ernst}{2021}]{b_and_e}
\begin{barticle}
\bauthor{\bsnm{Blechschmidt}, \binits{J.}},
\bauthor{\bsnm{Ernst}, \binits{O.G.}}:
\batitle{Three ways to solve partial differential equations with neural networks — a review}.
\bjtitle{GAMM-Mitteilungen}
\bvolume{44}(\bissue{2}),
\bfpage{202100006}
(\byear{2021})
\doiurl{10.1002/gamm.202100006}
{\href{https://arxiv.org/abs/https://onlinelibrary.wiley.com/doi/pdf/10.1002/gamm.202100006}{{https://onlinelibrary.wiley.com/doi/pdf/10.1002/gamm.202100006}}}
\end{barticle}
\endbibitem

\bibitem[\protect\citeauthoryear{Pang et~al.}{2019}]{pang_fpinn}
\begin{barticle}
\bauthor{\bsnm{Pang}, \binits{G.}},
\bauthor{\bsnm{Lu}, \binits{L.}},
\bauthor{\bsnm{Karniadakis}, \binits{G.E.}}:
\batitle{fpinns: Fractional physics-informed neural networks}.
\bjtitle{SIAM Journal on Scientific Computing}
\bvolume{41}(\bissue{4}),
\bfpage{2603}--\blpage{2626}
(\byear{2019})
\doiurl{10.1137/18M1229845}
\end{barticle}
\endbibitem

\bibitem[\protect\citeauthoryear{Lenau et~al.}{2025}]{lenau2024importance}
\begin{barticle}
\bauthor{\bsnm{Lenau}, \binits{A.}},
\bauthor{\bsnm{Dimiduk}, \binits{D.M.}},
\bauthor{\bsnm{Niezgoda}, \binits{S.R.}}:
\batitle{Importance of hyper-parameter optimization during training of physics-informed deep learning networks}.
\bjtitle{Integr Mater Manuf Innov}
\bvolume{14},
\bfpage{115}--\blpage{135}
(\byear{2025})
\doiurl{10.1007/s40192-025-00394-6}
\end{barticle}
\endbibitem

\bibitem[\protect\citeauthoryear{Isola et~al.}{2018}]{isola2018}
\begin{botherref}
\oauthor{\bsnm{Isola}, \binits{P.}},
\oauthor{\bsnm{Zhu}, \binits{J.-Y.}},
\oauthor{\bsnm{Zhou}, \binits{T.}},
\oauthor{\bsnm{Efros}, \binits{A.A.}}:
Image-to-Image Translation with Conditional Adversarial Networks
(2018)
\end{botherref}
\endbibitem

\bibitem[\protect\citeauthoryear{Efron and Tibshirani}{1994}]{efron1994boot}
\begin{bbook}
\bauthor{\bsnm{Efron}, \binits{B.}},
\bauthor{\bsnm{Tibshirani}, \binits{R.J.}}:
\bbtitle{An Introduction to the Bootstrap}.
\bpublisher{Chapman and Hall/CRC},
\blocation{New York}
(\byear{1994})
\end{bbook}
\endbibitem

\bibitem[\protect\citeauthoryear{Carslaw}{1930}]{GibbsPhen}
\begin{bbook}
\bauthor{\bsnm{Carslaw}, \binits{H.S.}}:
\bbtitle{Introduction to the Theory of Fourier's Series and Integrals (Third Ed.), Chapter IX}.
\bpublisher{Dover Publications Inc.},
\blocation{London}
(\byear{1930})
\end{bbook}
\endbibitem

\bibitem[\protect\citeauthoryear{Basri et~al.}{2019}]{Basri2019}
\begin{botherref}
\oauthor{\bsnm{Basri}, \binits{R.}},
\oauthor{\bsnm{Jacobs}, \binits{D.W.}},
\oauthor{\bsnm{Kasten}, \binits{Y.}},
\oauthor{\bsnm{Kritchman}, \binits{S.}}:
The convergence rate of neural networks for learned functions of different frequencies.
CoRR
\textbf{abs/1906.00425}
(2019)
{\href{https://arxiv.org/abs/1906.00425}{{1906.00425}}}
\end{botherref}
\endbibitem

\end{thebibliography}
\clearpage
\section{Supplemental Figures}
\subsection{Examples of training sessions reproducing Gibbs oscillations for each method}
\begin{figure}[H]
\centering
\captionsetup{width=\linewidth}
\includegraphics[width=0.8\textwidth]{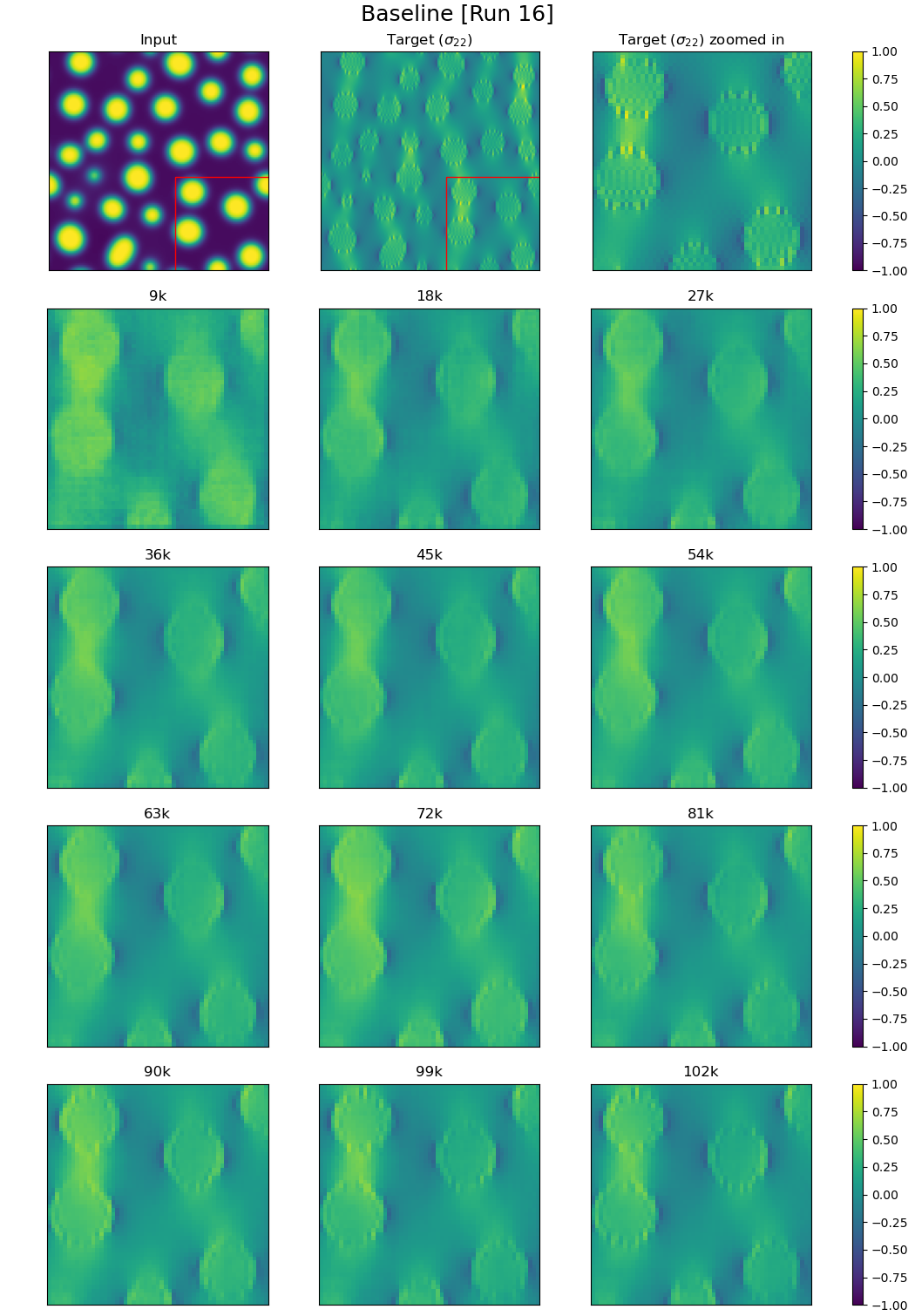}
\caption{Baseline predicted $\sigma_{22}$ field from different iterations saved throughout a single training. Replicates Gibbs oscillations.}
\label{gibbs b}
\end{figure}
\begin{figure}[H]
\centering
\captionsetup{width=\linewidth}
\includegraphics[width=0.8\textwidth]{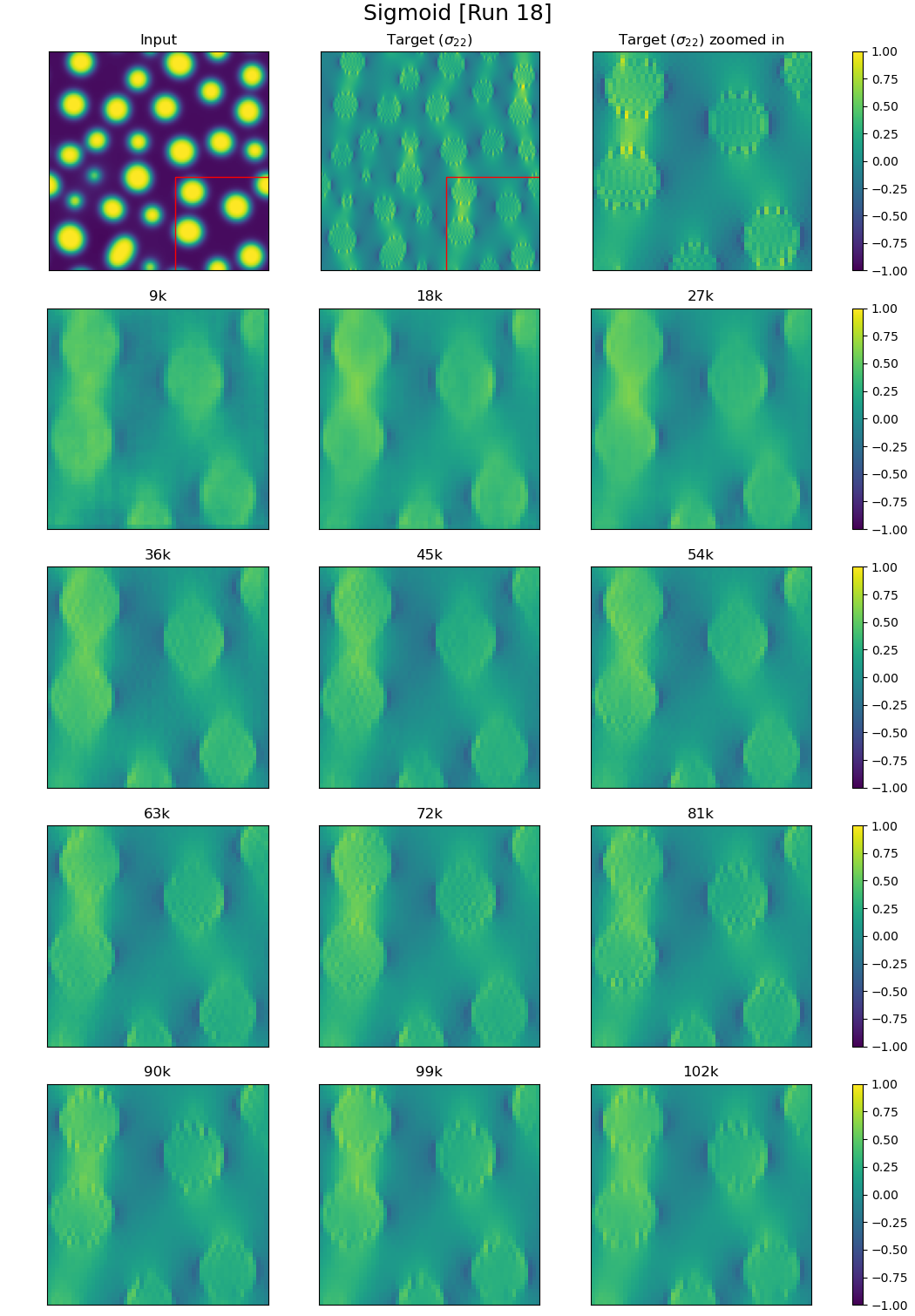}
\caption{Sigmoid predicted $\sigma_{22}$ field from different iterations saved throughout a single training. Replicates Gibbs oscillations.}
\label{gibbs si}
\end{figure}
\begin{figure}[H]
\centering
\captionsetup{width=\linewidth}
\includegraphics[width=0.8\textwidth]{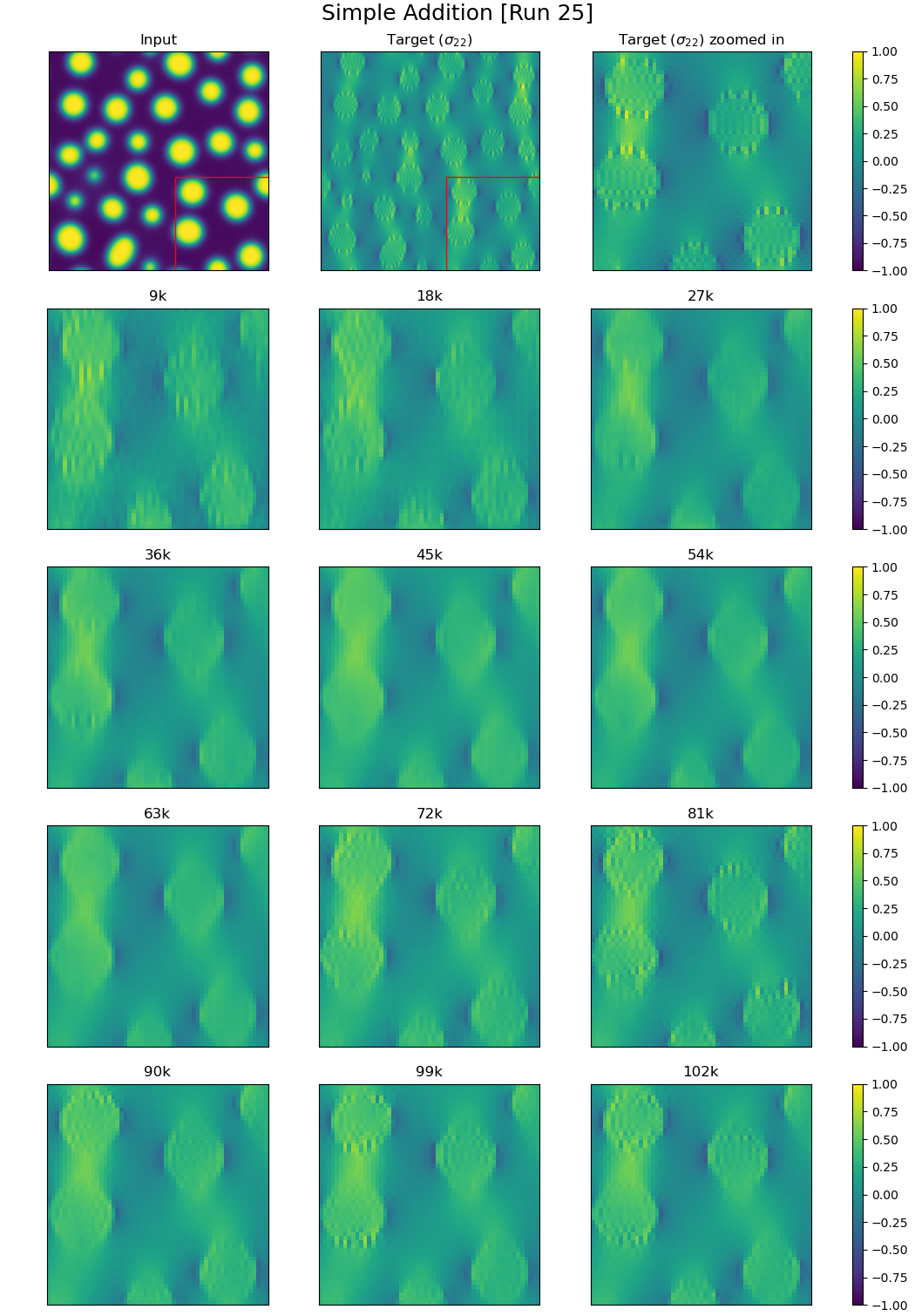}
\caption{Simple addition predicted $\sigma_{22}$ field from different iterations saved throughout a single training. Replicates Gibbs oscillations.}
\label{gibbs sa}
\end{figure}
\begin{figure}[H]
\centering
\captionsetup{width=\linewidth}
\includegraphics[width=0.8\textwidth]{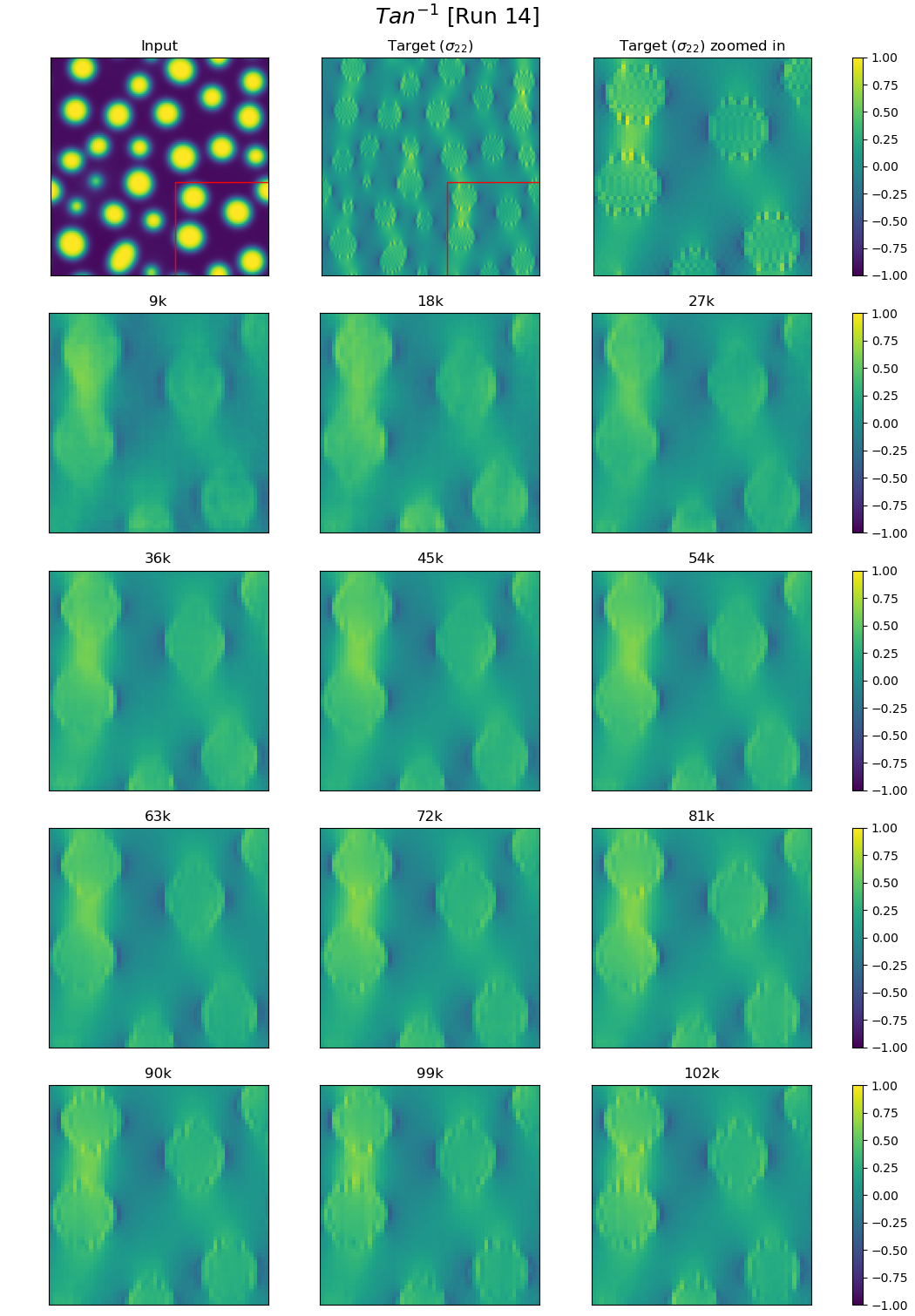}
\caption{$Tan^{-1}$ predicted $\sigma_{22}$ field from different iterations saved throughout a single training. Replicates Gibbs oscillations.}
\label{gibbs tn}
\end{figure}

\subsection{Examples of training sessions NOT reproducing Gibbs oscillations for each method}
\begin{figure}[H]
\centering
\captionsetup{width=\linewidth}
\includegraphics[width=0.8\textwidth]{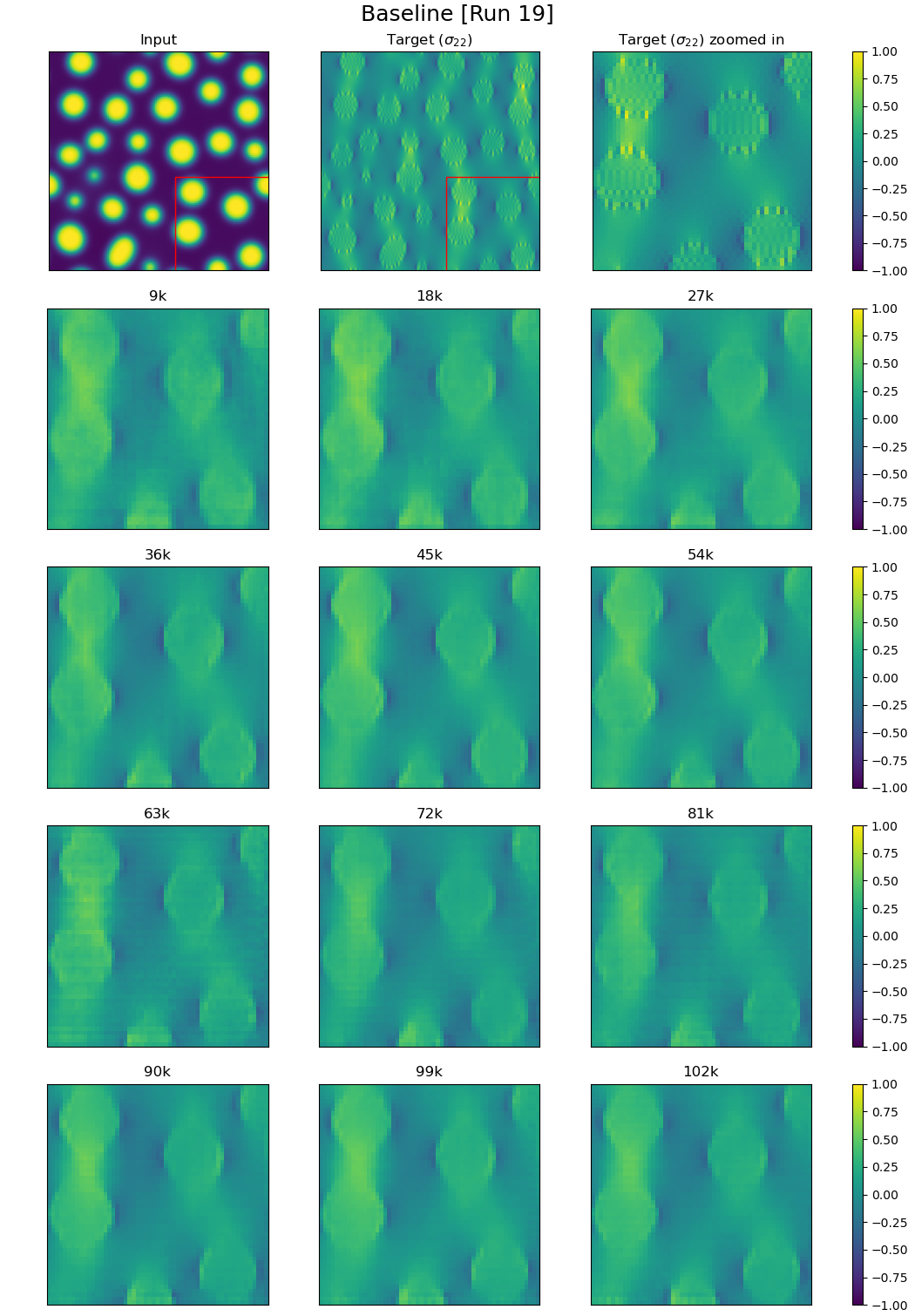}
\caption{Baseline predicted $\sigma_{22}$ field from different iterations saved throughout a single training. Does not replicate Gibbs oscillations.}
\label{nogibbs b}
\end{figure}
\begin{figure}[H]
\centering
\captionsetup{width=\linewidth}
\includegraphics[width=0.8\textwidth]{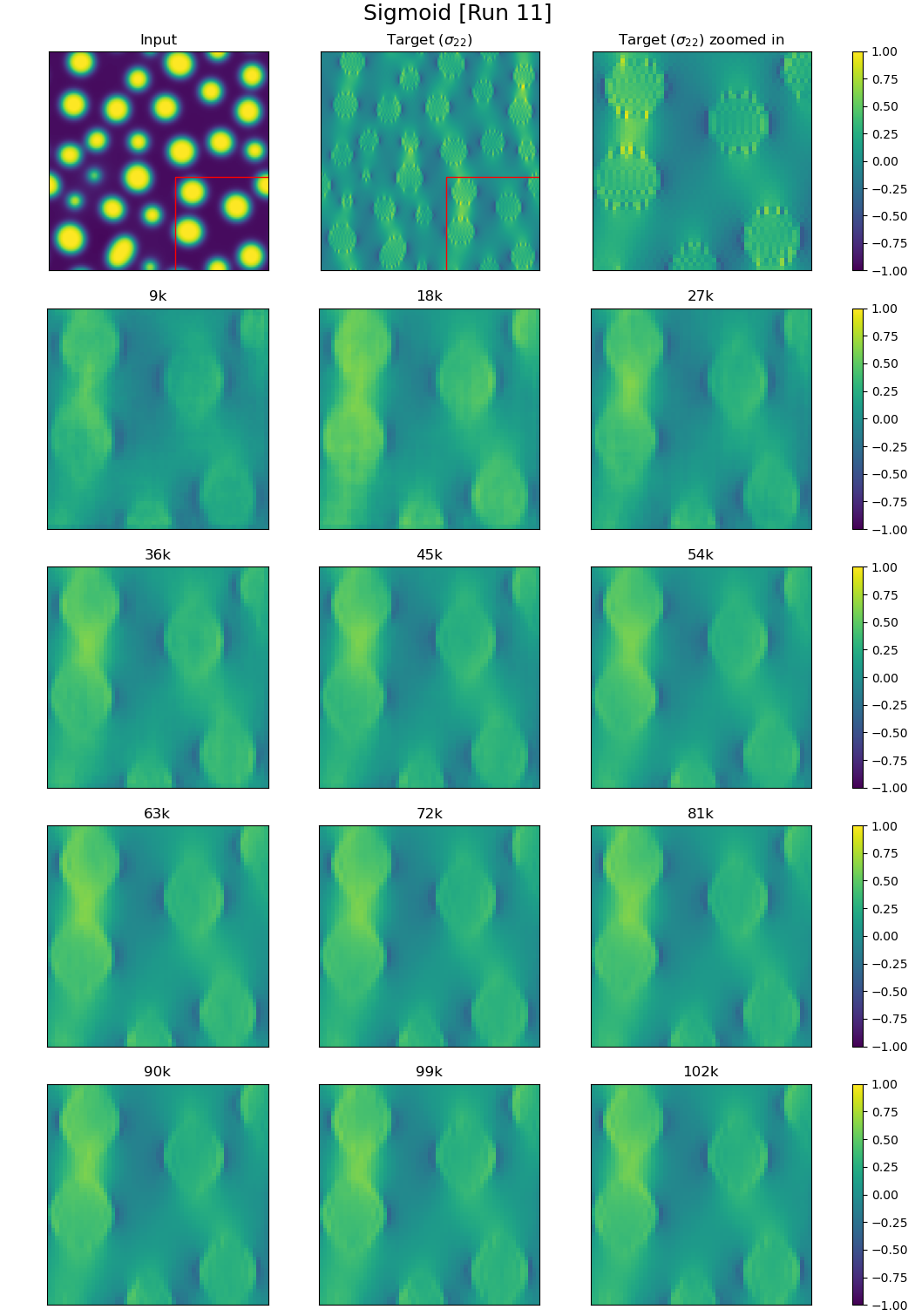}
\caption{Sigmoid predicted $\sigma_{22}$ field from different iterations saved throughout a single training. Does not replicate Gibbs oscillations.}
\label{nogibbs si}
\end{figure}
\begin{figure}[H]
\centering
\captionsetup{width=\linewidth}
\includegraphics[width=0.8\textwidth]{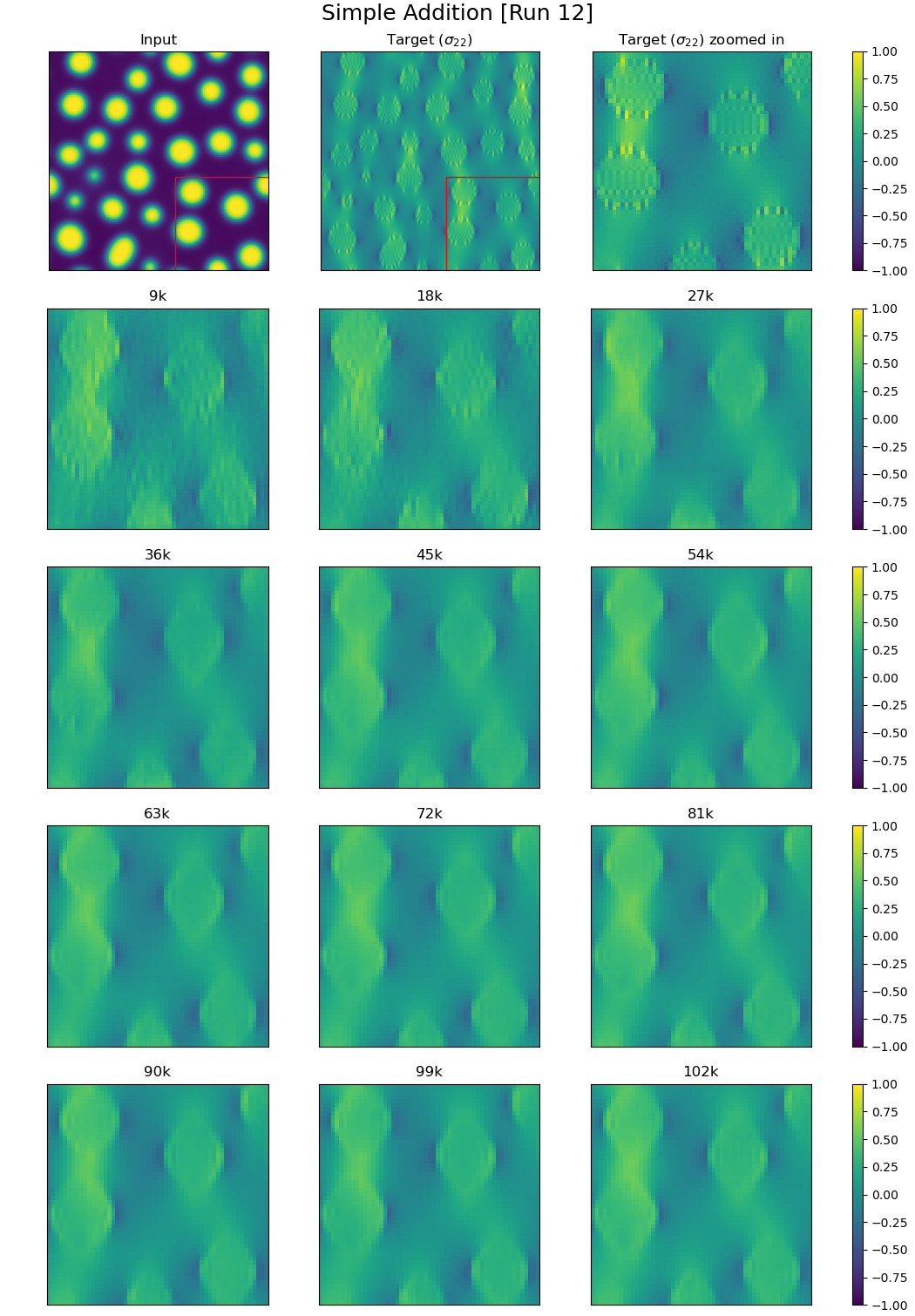}
\caption{Simple addition predicted $\sigma_{22}$ field from different iterations saved throughout a single training. Does not replicate Gibbs oscillations.}
\label{nogibbs sa}
\end{figure}
\begin{figure}[H]
\centering
\captionsetup{width=\linewidth}
\includegraphics[width=0.8\textwidth]{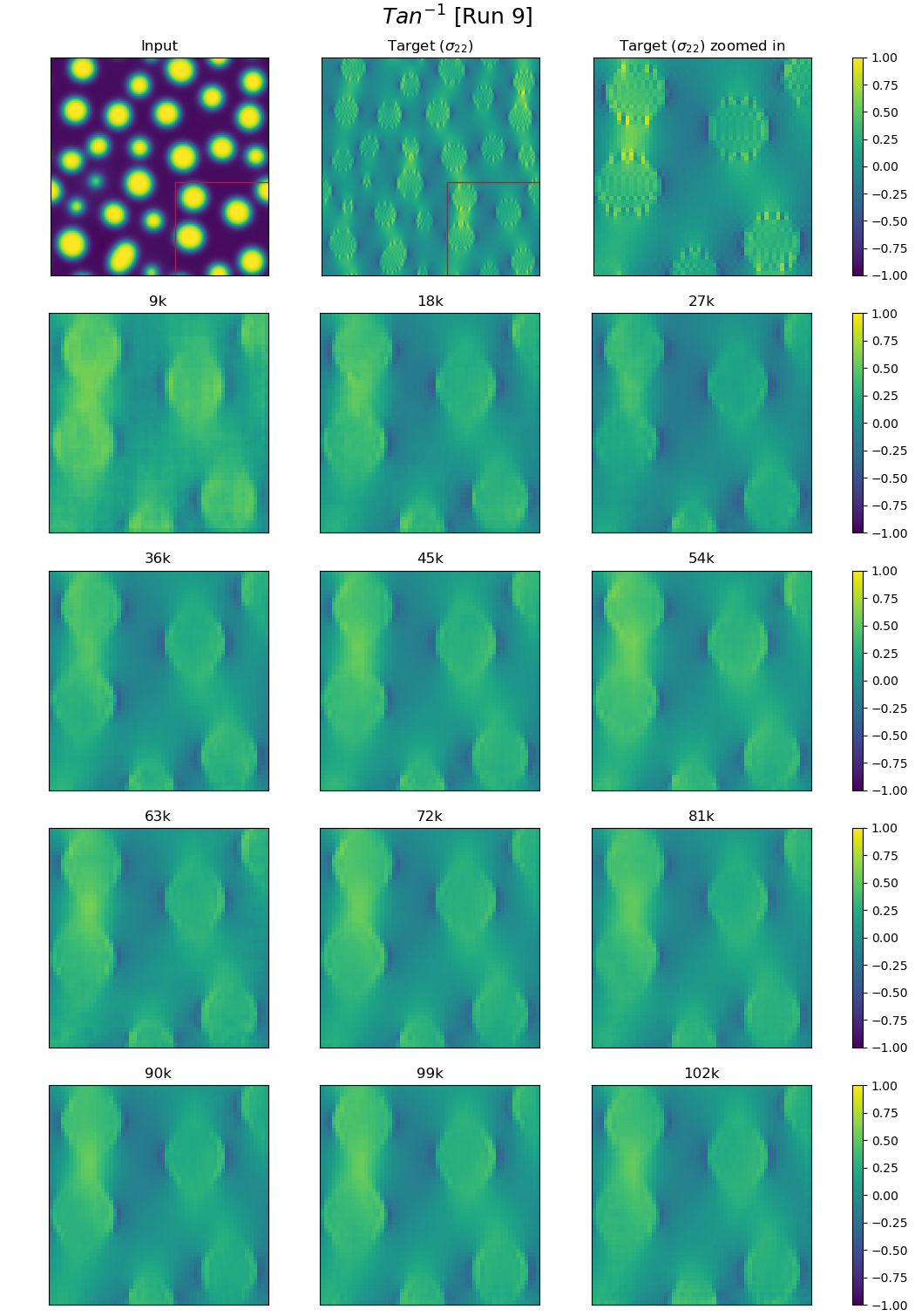}
\caption{$Tan^{-1}$ predicted $\sigma_{22}$ field from different iterations saved throughout a single training. Does not replicate Gibbs oscillations.}
\label{nogibbs tn}
\end{figure}

\end{document}